\pdfoutput=1

\documentclass[fleqn]{article}

\usepackage{arxiv}

\usepackage{amssymb}

\usepackage{xcolor} 

\usepackage[utf8]{inputenc} 
\usepackage[T1]{fontenc}    
\usepackage{hyperref}       
\usepackage{url}            
\usepackage{booktabs}       
\usepackage{amsfonts}       
\usepackage{microtype}      
\usepackage{graphicx}

\usepackage{subfig}  

\usepackage{amsmath} 

\usepackage{amsthm}

\usepackage{booktabs} 
\usepackage{etoolbox}
\usepackage{xspace}
\usepackage{tabularx}
\usepackage{fancyvrb}
\usepackage{letltxmacro}

\renewcommand{\sec}{Sect. } 
\newcommand{\fig}{Fig. }

\newcommand{\up}{\mbox{UP}} %

\newcommand{\down}{\mbox{DOWN}} %

\newcommand{\rt}{\mbox{RIGHT}} %
\newcommand{\rtb}{\mbox{RIGHT }} %

\newcommand{\lft}{\mbox{LEFT}} %
\newcommand{\lftb}{\mbox{LEFT }} %

\newcommand{\food}{\mbox{FOOD}} %
\newcommand{\foodb}{\mbox{FOOD }} %

\newcommand{\empt}{\mbox{EMPTY}} %
\renewcommand{\empty}{\mbox{EMPTY}} 
\newcommand{\emptyb}{\mbox{EMPTY }} 

\newcommand{\barr}{\mbox{BARRIER}} %
\newcommand{\barrb}{\mbox{BARRIER }} %

\newcommand{\tx}{\tilde{x}} 
\newcommand{\ty}{\tilde{y}} 

\newcommand{\fn}{function}
\newcommand{\fns}{functions}
\newcommand{\fnsb}{functions }

\newcommand{\pmapb}{\mbox{ProbMap }} %
\newcommand{\pmap}{\mbox{ProbMap}} %
\newcommand{\probmapb}{\mbox{ProbMap }} %
\newcommand{\probmap}{\mbox{ProbMap}} %

\newcommand{\unvisited}{\mbox{LeastVisited }} %
\newcommand{\unvisitedn}{\mbox{LeastVisited}} %

\newcommand{\oracle}{\mbox{Oracle }} %
\newcommand{\oraclen}{\mbox{Oracle}} %

\newcommand{\random}{\mbox{Random}} %
\newcommand{\rand}{\mbox{Random}} %
\newcommand{\randb}{\mbox{Random }} %

\newcommand{\biased}{\mbox{Biased}} %
\newcommand{\biasedb}{\mbox{Biased }} %

\newcommand{\greedy}{\mbox{Greedy}} %
\newcommand{\greedyb}{\mbox{Greedy }} %

\newcommand{\pathmem}{\mbox{Path }} %
\renewcommand{\path}{\mbox{Path}} 
\newcommand{\pathb}{\mbox{Path }} 

\newcommand{\done}[1]{}

\newtheorem*{theorem*}{Theorem} 

\newtheorem*{lemma*}{Lemma} 

\newtheorem*{corollary*}{Corollary}

\newcommand{\p}{p}  

\newcommand{\mysim}{\raise.17ex\hbox{$\scriptstyle\sim$}}
\newcommand{\mytilde}{\raise.17ex\hbox{$\scriptstyle\sim$}}

\newcommand{\ie}{{\em i.e.~}} 
\newcommand{\vs}{{\em vs.~}}  
\newcommand{\eg}{{\em e.g.~}} 
\newcommand{\etal}{{\em et~al.~}}

\newcommand{\co}[1]{}

\title{When Remembering and Planning are Worth it: Navigating under
  Change}

\date{} 					

\author{Omid Madani\thanks{Correspondence to omidmadani@yahoo.com. The
    authors were partially affiliated with Brown University.} \and J. Brian Burns \and Reza Eghbali
  \and Thomas L. Dean}

\renewcommand{\headeright}{}
\renewcommand{\undertitle}{}

\hypersetup{
pdftitle={Memory-Based Navigation Strategies in Changing Worlds}
pdfsubject={} 
pdfauthor= {Omid Madani}
pdfkeywords={Navigation, Non-Stationarity, Memory, Learning, Planning, Agents }
}

\begin{document}
\maketitle

\begin{abstract}



We explore how different types and uses of memory can aid spatial
navigation in changing uncertain environments.  In the simple foraging
task we study, every day, our agent has to find its way from its home,
through barriers, to food.
%
Moreover, the world is non-stationary:
from day to day, the location of the barriers and food may change,
and the agent's sensing such as its
location information is uncertain and very limited.
%
Any model construction, such as a map,
%
%
and use, such as planning, needs to be robust against
these
challenges,
%
%
and if any learning is to be useful, it needs to be
adequately fast.
We look at a
range of strategies, from simple to sophisticated, with various uses
of memory and learning.
%
We find that an architecture that can incorporate multiple strategies
is required to handle (sub)tasks of a different nature, in particular
for exploration and search, when food location is not known, and for
planning a good path to a remembered (likely) food location.  An agent
 that utilizes non-stationary probability
learning techniques to keep updating its (episodic) memories and that
uses those memories to build maps and plan on the fly (imperfect maps, \ie
noisy and limited to the agent's experience)
%
%
can be
increasingly and substantially more efficient than the simpler
(minimal-memory) agents, as the
task difficulties such as distance to goal are raised, as long as the
uncertainty, from localization and change, is not too large.
\end{abstract}

\vspace*{.25in}


\hspace*{0.0in} \mbox{\em "You can never know everything," Lan said
  quietly, "and part of what you know is always wrong. Perhaps even}\\
\hspace*{0.07in}\mbox{\em the most important part. A portion of wisdom lies
  in knowing that. A portion of courage lies in going on
  anyway."}\footnote{Excerpt are from ``The Bayesian Choice'',
    by C. Robert \cite{choice18} (thanks to Kevin Murphy for the book gift).}\vspace*{.1cm}
 \hspace*{2.5in}\mbox{Winter's Heart, Book IX of the Wheel of Time, by
   Robert Jordan.} \\

\hspace*{0.03in} \mbox{\em Alice remarked, ``I can't remember things before they happen.''}\\
\hspace*{0.041in}\mbox{\em``It's a poor sort of memory that only works backwards,'' says the White
  Queen to Alice. }\footnote{We saw this first in G. Buzsaki's book, ``The Brain
    from Inside Out'' \cite{insideout19}.}\vspace*{.1cm}
  \hspace*{3.3in}\mbox{Through the Looking Glass, by Lewis
    Carroll.}

  \keywords{Spatial Navigation, Foraging, Non-Stationarity, Autonomous
    Agents, Memory, Sample Efficiency, Sample Bias, Continual Learning
    and Planning, Planning under Uncertainty, Reinforcement Learning,
    Learning in a Lifetime}



\co{

}


\section{Introduction}
\label{sec:intro}





%
%

The
rich, productive, and
ever changing world necessitates agents capable of continuous
and flexible adaptations
to achieve their objectives.  In the world of engineered AI systems,
reinforcement learning (RL) techniques \cite{rl1,rl2}, specially the
model-free variety using deep feed-forward neural networks, have had
substantial success in the past decade, as they are flexible, in that
they do not assume much about the world, \eg do not require modeling
and encoding a complex world by the  engineers of the
agent, and powerful, as the (policy or value) function to be learned
can be highly complex
\cite{Silver2016MasteringTG,Christiano2017DeepRL,mnih2013}.
%
%
%
%
However, a
critical limitation of the current model-free RL is the requirement
of vast amounts of data.
In dynamic real-world settings where the agent must adapt to evolving
and changing tasks—ranging from minor adjustments to fundamental
shifts—such systems often fall short: a change in the function (task)
to be learned or performed can effectively reduce the available training
sample size.
Extensive and expensive simulations and substantial pretraining are
common workarounds,
striving to anticipate and train for {\em all possibilities}, as a way
to enhance robustness, but the demands of the real productive world
often curbs the success of such approaches.
We seek efficient online learning and continual adaption that occurs
{\bf \em in a lifetime, for a lifetime} (\ie  keeping pace with
life's changes and challenges).

Model-based techniques can potentially address some of the drawbacks
of data inefficiency,
their promise being that once
acquired, the model(s) can be
used repeatedly
to find solutions for a diversity of related tasks. A challenge is
what exactly an explicit model could mean, and whether such a model
can be efficiently learned and updated.  Consider spatial navigation,
\eg for foraging, which is a fundamental activity across the
biological spectrum, with much research exploring how different minds
meet
its challenges
\cite{Zaburdaev2014LevyW,forage97,wehner1997ant,antsPlos,fly_nav1,innate2022,Tolman1948,connor19,ingold2000}.
\co{
Choosing a movement action
presents an interesting special case of the general problem of action
selection:
upon execution, the agent's location or orientation may change within
the environment, but
otherwise the environment itself is not altered (directly): here, the
stakes are not so high as general decision making scenarios, since,
often, one can undo one's movement action. On the other hand, }
Organisms need to be efficient (in energy/time) and flexible, and
perform different but related navigation task (get to food, water,
shelter, ...) in an at times dangerous and changing terrain (seasonal
changes as well as abrupt changes, such as floods and droughts).  The
reward, from reaching the goal, is often distant, and the goal can
change, thus the slow learning via reward propagation is often
insufficient.
From bacteria to bats and birds, living organisms utilize a range of
sensing, memory, communication, and computational (e.g., inference)
capabilities to efficiently reach their destinations
\cite{ants2011,Tolman1948,connor19}.\footnote{Organisms of the same
species, and the same organism but at different times, can use
different (mix of) strategies \cite{Wijnen2024RodentMS,Igloi2009SequentialES,Lee2024StochasticCO}.
The same person could use several strategies to get to a
destination, \eg from deciphering signs on a subway map to asking
other people for directions (and remembering and executing those rough
directions). } While
much of this machinery appears innate \cite{innate2022}, a significant
portion is likely dynamically learned and repeatedly
tuned and reconfigured throughout an organism's lifespan.
In particular, the hippocampus is a structure that
is established to be critically involved in memory formation and use,
for instance to help the agent navigate via the creation of the
so-called {\em
  cognitive maps} (such as place cells and grid cells) 
\cite{okeefe78,Tolman1948}, though the details,
such as what is represented and how such is used, continues to be
debated and investigated
\cite{Yaghoubi2024PredictiveCodingReward,structuringKW2020,FIEDLER2021,Stachenfeld2017theHippo}.


A map data structure, once built, is versatile and highly useful for
navigation:
%
at least theoretically, to get from any point A to any point B (on the
map), one can plan using the map and execute the plan, \ie one can
solve a range of (related) navigation tasks rather efficiently when
one has access to a good map. The map can be modified and reused if
reroutings are sought (\eg in case of new barriers blocking the routes
that used to work). Consequently maps are
the go to data structures for navigation tasks, \eg in robotics and in
particular SLAM domains \cite{slam2016}. But building a model (map) of
a complex changing world, under limited time and sensing, carries its
own
many challenges.  We explore these challenges in a simplified world
and task, akin to the animats work \cite{wilson1985KnowledgeGI,Strannegrd2018LearningAD}: Consider a simple
grid-world where every day an agent, with very limited sensing of its
world, needs to
find its way to food through barriers (road blocks). Moreover, the
routes to food can change from day to day, some times substantially:
several barriers or food may change location. A map could
be part of a solution.
If the world is fairly static, such a map could save much time over
the lifetime of the agent. There are a number of challenges,
including:
\begin{enumerate}
\item Any map learnt will be biased
towards the experience of the agent, for instance, how much
exploration it has performed (biased non-IID samples).
\item The world changes, and information extracted from the map can be
  out-dated (uncertainty 1).
\item Agent's knowledge
of its current location (we use simple path integration) contains
errors due to motion (action) noise (uncertainty 2)
\item How (and whether) an agent would carve and granularize its
  sensed and perceived space into locations, to serve its needs,
  remains open. 
\end{enumerate}
We
explore
the first three challenges in this work.
Issues of world complexity and substantial uncertainty has thwarted
the practical use of model-based techniques for open-ended real-world
tasks that contain a diversity of uncertainties. Probabilistic
planning is highly intractable
in general \cite{Madani2003OnUnd,Littman1998TheCC}, and earlier
works on agents
have also found that the focus on explicit representations and
planning in traditional AI approaches may be misplaced, in part to due
to the aforementioned challenges, and in part because simpler agent
strategies can be sufficiently successful in
a diversity of worlds and tasks
\cite{Agre1987PengiAI,Agre1990WhatAP,Brooks1991IntelligenceWR}.

In this paper, we investigate a number of
navigation strategies, from simple to
the more sophisticated, to see how they
compare as we
vary certain aspects of task difficulty: the
environment size (distance to food), the proportion of barriers (path
complexity), and two types of uncertainty: daily barrier/food location
change and the uncertainty of localization (in agent location).  In
particular, each agent type can use a mix of (pure) strategies to get to
food, and we compare the more sophisticated agents to a fairly simple
mixed greedy agent (Table \ref{tab:steps}), that uses random action
selection some of the time, together with the strategy of (greedily)
lowering its distance to goal, at other times. A fly, for example, may
execute a strategy akin to greedy via the use of the smell sense
\cite{fly2,fly_nav1}.

In general, the simpler strategies require less (of memory and
computing machinery)\footnote{Simpler strategies may rely on more
sophisticated sensing: sensing itself can be compute
intensive and involve pipelines of processing as well, for instance
for estimating any of localization, orientation, and wind direction
\eg for smell \cite{Heinze2018PrinciplesOI,fly2,fly_nav1}.}  and can
remain useful in a wider range of environments: In environments where
food is abundant and near, and obstacles are sparse, they would be
adequate.  However, in harsher more challenging and richer
environments,\footnote{One view in philosophy of biology posits that
the organism (agent) itself, with its capabilities and interactions,
determines its own environment \cite{maturana2002AutopoiesisSC}.}  and
when there exists (sufficiently stable) structure to the world, and
enough time to make a decision (think, reason, ...), the more
sophisticated strategies may outperform the simpler ones.  We have two
goals in this work:
\begin{itemize}
\item We ask: {\em Are the more sophisticated map-based strategies
  worth their costs (of memory, intricate control, and in general
  compute machinery)?} Under what conditions are they better than the
  simpler ones?  By how much?
\item (upon a positive answer wrt their worth) We provide insights on
  agent architecture and the types of memory and learning
  that could support efficient map construction and
  effective updates (map maintenance), and map use.
\end{itemize}


We find that well-designed memory-based strategies, that appropriately
take uncertainties into account, in building, updating, and using
memories that ultimately serve as maps (in this paper), outperform
mixed-greedy and other simple strategies, and the advantage grows, \eg
to over 20x fewer steps to food, with environment size and difficulty
(distance to goal and barrier portion), as long as the rate of change
and location-uncertainty is not too large.  Due to uncertainty and
task complexity, pure strategies, such as always planning for a goal,
may underperform drastically, and we find that robust behavior needs
to use several pure strategies to handle (sub)tasks of a different
nature: search and exploration, when food location is unknown, and
planning a good path or moving towards food when (likely) food
location is known (\sec \ref{sec:tasks}).  We describe an architecture
where the agent uses one (active) strategy at any given time, but
communicates the action taken and the sensory input to all strategies
(so that memory-based strategies can update their memories
appropriately).
Furthermore, a planning-based strategy needs to take failure at
planning time (\eg no path to food) or execution failure (eg an
unanticipated barrier) into account: Repeated (re)planning, as well as
memory (map) updating, are necessities. Thus, there is indeed
complexity to appropriate implementation of the more sophisticated
strategies, but the gains in flexibility and reach can be worth it.

How does an agent (come to) know what to remember and how best to
utilize its (episodic) memory?  We assume certain capabilities, such
as basic sensing, the importance of space and the details of
path-integration, are (mostly) hard-wired
\cite{Starck1998AvianGA,Estes1991TheBG,innate2022}.
In future work, we hope to reduce the number and the extent of the
'hard-wired assumptions' and in particular add additional learning in
a lifetime.

This paper is organized as follows: We describe the simple grid
environment and the basics of the agent and task(s) next.
We then describe our flexible agent structure, which allows for
incorporating and interfacing with multiple strategies (\sec
\ref{sec:agents}), followed by describing the (pure) strategies, with
different uses of sensing, memory, learning, and plannings, in \sec
\ref{sec:sts}. The code is publicly available at GitHub
\cite{barriers_code}.  \sec \ref{sec:exps} presents our
experiments, and \sec \ref{sec:related} discusses related work.
We conclude
in \sec \ref{sec:summary}. A shorter version of this work appeared in
BICA \cite{flatland2}.


\section{The Environment, the Agent, and Task(s)} 
\label{sec:env}

\begin{figure}[tb]
\begin{center}
\centering
\begin{minipage}{\linewidth}
  \subfloat{{\includegraphics[height=2.5cm,width=16cm]{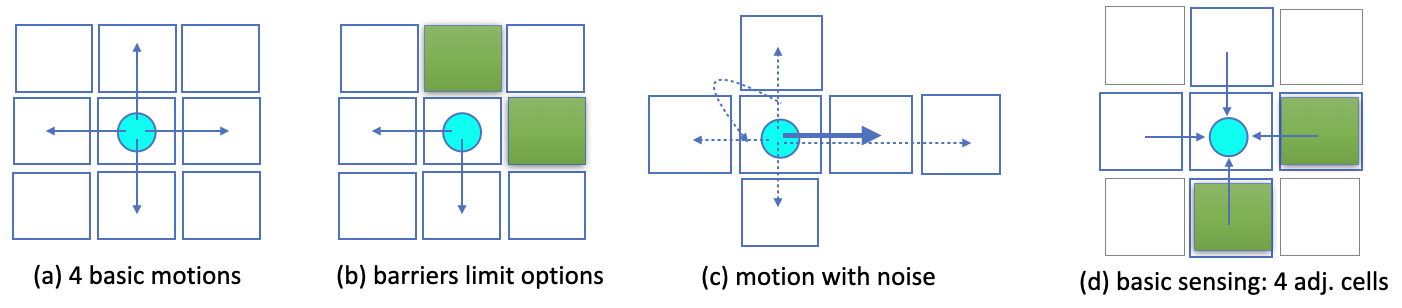}}  }
\end{minipage}
\end{center}
\vspace{.2cm}
\caption{Basic actions and sensing: (a) 4 possible actions: \lft (west), \rt, \up, or \down. (b) In
  this example, with two barriers, the agent has two legal actions
  (left and down). (c) Motion noise, up to 6 possibilities: when
  intending to go east (right), with some (noise) probability, the
  agent may end up in another location: stay in the same cell, go up,
  or down, or left, or go two hops east. (d) Sensing is also from
  a single adjacent cell (4 such).   }
\label{fig:motion_and_sensing} %
\end{figure}

Our environment is an $N$ x $N$ grid of $N^2$ cells, each cell
identified by its location coordinates $(x, y)$.  There is a single agent, with limited
sensing (to be described). The agent is in exactly one cell of the
grid
at any time, and can execute a (move) action to change its location by
one cell.
Time is broken into day and time tick: day $1, 2, 3, \cdots$, and
within a day, time ticks
 $t=1, 2, \cdots$. 
 We focus on a simple closed-world:
 a cell is in one of three states at any time:
 {\bf \empt},  {\bf \barr}, or {\bf \food}.  The agent has four
actions: move to a single adjacent cell without \barrb ({\bf \up},
{\bf \down}, {\bf \lft} or {\bf \rt}).
With
{\bf \em motion-noise probability} $p$,
for a low $\p \in [0,1]$ (\eg $p=0.02$)
the environment picks an
alternative 
'noisy' position, and this includes staying in the
same location and moving two steps forward
(\fig \ref{fig:motion_and_sensing}(b)).
%
%
An {\bf illegal} action is one leading to a barrier and has no effect.
The agent cannot go off the grid (assume barriers).  Upon action
execution, time tick is incremented.


\begin{figure}[tb]
\begin{center}
\begin{minipage}{0.4\linewidth}
 \hspace*{-1.5cm}  \subfloat{{\includegraphics[height=4cm,width=5cm]{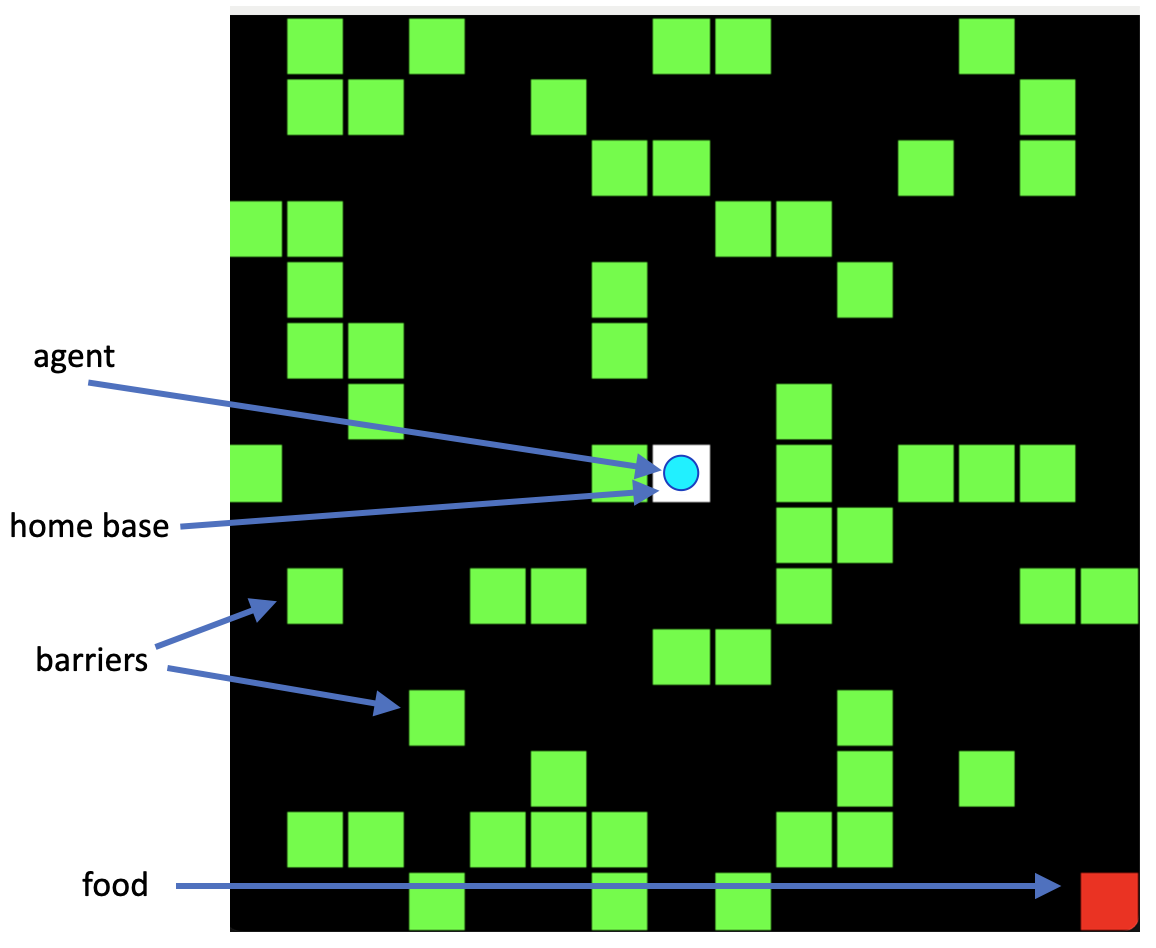}}  }\\ \\
      \hspace*{0.35cm} \small{ \mbox{\ \ \ \ \ (a) The task.}  }
\end{minipage}
\hspace*{-.13in}\begin{minipage}{0.35\linewidth}
  \subfloat{{\includegraphics[height=2cm,width=4cm]{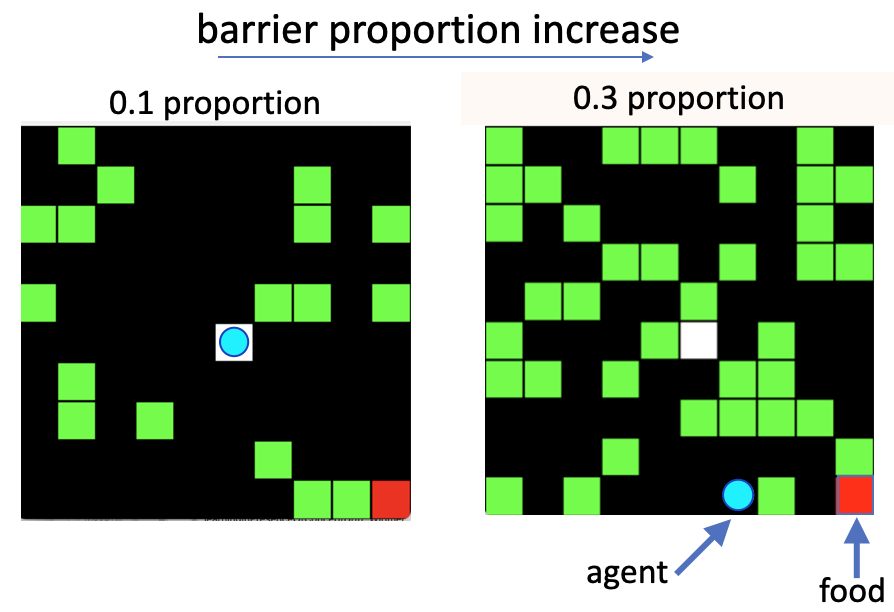} }}\\
  \subfloat{{\includegraphics[height=2cm,width=4cm]{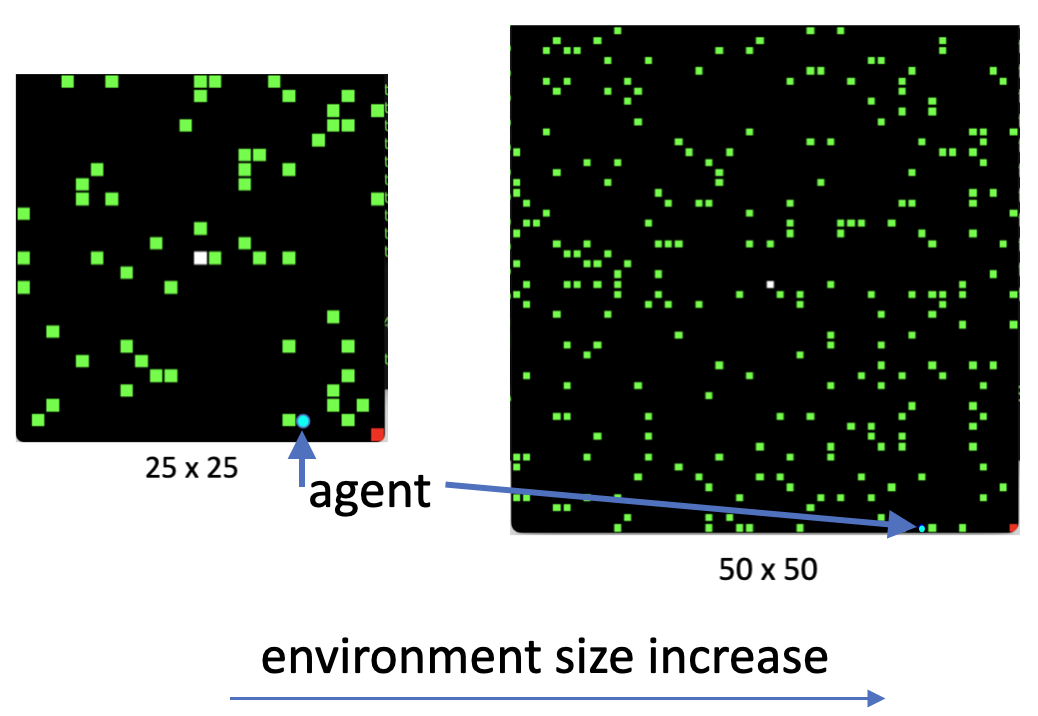} }}\\
      \small{         (b) 2 ways of raising task difficulty.  }
\end{minipage}
\hspace*{.11in}\begin{minipage}{0.2\linewidth}
  \subfloat{{\includegraphics[height=3cm,width=2cm]{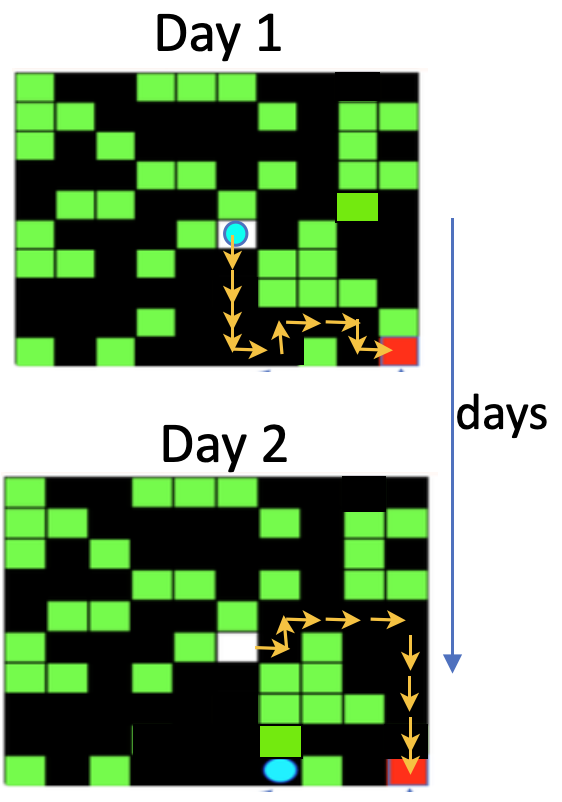} }} \\
      \small{         (c) Day to day food routes can change.  }
\end{minipage}

\end{center}
\vspace{.2cm}
\caption{(a) The agent and its environment. It is important to
  emphasize that the agent {\em does not see} the whole grid, just the
  locations immediately adjacent to its current location (partial
  observability). (b) Two knobs on task complexity: barrier proportion
  and environment size (distance to goal). c) A 3rd knob on task
  difficulty: rate of (barrier) change, from day to day.  }
\label{fig:envs} %
\end{figure}

Activity in the environment is broken into daily trips: in our
experiments, each day, the agent starts at the home base, $(0, 0)$,
and the task of the agent is to get to the food location,
and save steps in doing so.\footnote{In this work, we ignore costs of
computation (in particular planning) costs (time or energy), and
assume any computation by the agent is performed within the budgeted
time/computation bounds.  }
The state of a cell is not changed within a day,
and there is always exactly one food cell.\footnote{One cell with food
can reflect a region that has food in a more realistic setting.}, and
we ensure that the generated environments are such that a path to food
exists.
Once the food is reached, the next day begins
and the agent location is reset to home $s=(0, 0)$.
%
The food location does not change often, \ie the food is often
replenished, from day to day (so the strategy of trying to go back to
the same place food was found tends to be useful). In most
experiments, from day to day, we change several barrier locations, but
keep the food in the same location (lower right corner).



\subsection{The Task(s): Getting to Food}
\label{sec:tasks}


A memory based agent deals with two tasks of a different nature in
getting to food.  Initially, on day 1, the agents doesn't know where
the food is, nor the size of the grid, etc. (see the next section on
limited sensing).\footnote{However, the strategies can be viewed as
being designed for or having certain implicit/encoded assumptions in
order to succeed in this type of task, such as 'it is useful to keep memories of
what was observed at each location'.}  On the first day, and if
(whenever) the food location changes, the task is more of a {\bf \em
  Search} problem, \ie search {\em in the environment}, for where food is,
and strategies geared toward
exploration are more successful. For an agent that doesn't
have
(episodic) memories, the task will always remain a
search-in-the-environment problem.  However, under some level of
stability in the environment/task,
agent could remember certain aspects during each day to help it
better navigate and reach food in that day and in future days.  Thus in
subsequent days, the task may become more of
a path {\em \bf Planning} problem, \ie search, but {\em internally},
using one's memories, for a good path (internal search).
But this is only if the agent can remember the relevant aspects.  Each
agent type we experiment with uses a different mix of (one or more) basic strategies, and
different strategies involve different types of memory (\eg short
within-day and long-term) and sensing (\sec \ref{sec:agents}).

\subsection{Change}
\label{sec:change}

From one day to next, several barriers may disappear and some new ones
may appear. In our experiments, we use the (barrier) {\em \bf
  change-rate} to set this change: a change-rate of 0.1 means that
about 10\% of the previous day's barriers are removed, and a similar
number of new ones are added (overall barrier proportion kept the same
from day to day).  In one set of experiments, we also change the food
location (\sec \ref{sec:food_change}).



\subsection{Limited Sensing (Observing, Localizing, ..) }
\label{sec:sense}


Animals use a variety of
sense modalities for navigation, such as hearing (echo-location of
bats), kinesthesis, olfactory, and vision.  In our work, we support a
few basic sense capabilities and different strategies may use a subset
of them.
One available sense
is looking one cell adjacent/around to get
its state (\food, \barr, \empt)
(a visual radius of 1).
For the greedy strategy, we assume the agent has a sense akin to
smell,
telling the agent which of the 4 actions reduces distance to
goal.\footnote{We do not model noise in the greedy/smell direction (and one
could increase such noise as the distance to food grows). On the other
hand, smell can be more powerful and yield more information, such as
the rough distance to goal, and in an extreme,
one can imagine the barriers as impenetrable and tall, and the odor's rout to
the agent translating to a path to the food.

}
All strategies know which
actions are legal.
%



The more sophisticated strategies require {\em \bf localization}:
access to
an estimate $(\tx, \ty)$ of
the true current location $(x,
y)$.
Our agent performs simple
{\bf path integration} from its home $(0,0)$,
keeping two counters (sums), one for the horizontal, another for
the vertical dimension.
A \rtb
increments the horizontal counter, a \lftb decrements it
(\eg $(0,0)$ becomes $(-1,0)$), and so on.



\co{
Simulating motion (localization) uncertainty: The agent submits its
selected action to the environment to be executed but with
{\bf \em motion-noise probability} $p$,
for a low $\p \in [0,1]$ (\eg $p=0.02$)
the {\em noisy-motion case} occurs: the environment picks an
alternative action or noisy position, and this includes staying in the
same location (\fig \ref{fig:motion_and_sensing}(b)).  In the
noisy-motion case, a position is picked as follows:
In some fraction $p_2$ of the noisy-motion cases, (by default $1/2$ of
such cases, \ie $p_2=0.5$), the location picked is either two steps
forward (in the same direction of the original action) or staying in
the same place (the two outcomes equally likely). And $1-p_2$ of the
noisy-motion cases, a position is picked uniformly at random from all
the available adjacent possibilities (up to 4).
}

Note that with a positive motion-noise $p$, the inaccuracy of the
path-integration estimate $(\tx, \ty)$ during the day is expected to
grow with time tick $t$ and in general the farther the agent is from
its home base (starting point).


\vspace*{.21cm} {\bf Dimensions of Difficulty.} In some experiments we
change the barrier portion or the grid size to change the difficulty
of the task. For instance, a higher barrier portion means longer and
more intricate paths to food, and remembering where barriers are or
the successful past paths can become more useful. On the other hand,
increasing the uncertainty, in our setting the barrier change-rate or
the motion noise, can counteract the benefits of remembering.


\section{Agent Structure} 
\label{sec:agents}



\begin{figure}[t]
  \begin{minipage}[t]{0.20\linewidth}
    \subfloat{{\includegraphics[height=2.5cm,width=5cm]{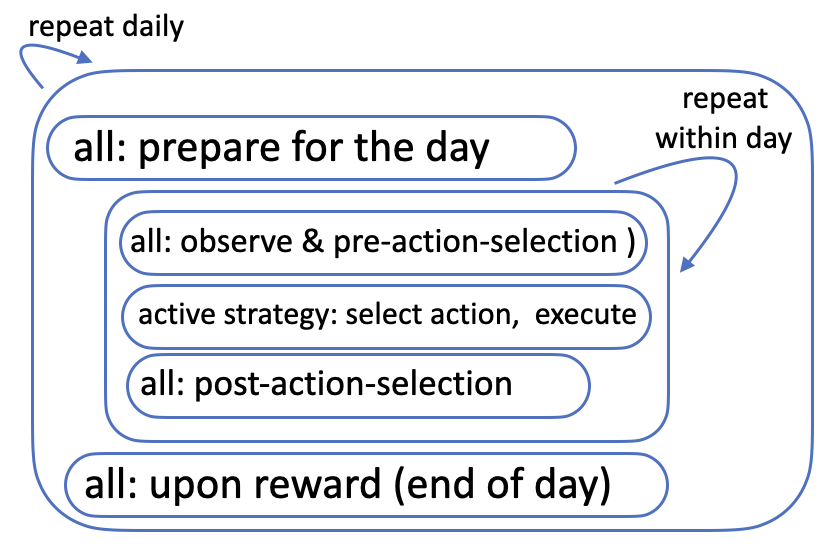} }}
  \end{minipage}
  \hspace*{0.90in}\begin{minipage}[t]{0.70\linewidth}
  Repeat (every day): // an agent's daily activity. \\
  \hspace*{0.3cm} Prepare for a new day: for each strategy, invoke its {\bf new-day \fn}. \\
  \hspace*{0.3cm}  Repeat until reward (\ie food is reached) \\
  \hspace*{0.8cm}  Get observations (interface with the environment) \\
  \hspace*{0.8cm}  For each strategy, invoke its {\bf pre-action-selection \fn} \\
  \hspace*{0.8cm}  Select an action using the active strategy ({\bf action-selection \fn}) \\
  \hspace*{1.1cm}  If failure or times up, change strategy (round-robin \& time budgets) \\
  \hspace*{0.8cm}  For each strategy: invoke {\bf post-action-selection \fn} \\
  \hspace*{0.8cm}  Execute action (interface with the environment) \\
  \hspace*{0.3cm} End of day: for each strategy, invoke its {\bf upon-reward \fn} \\
\end{minipage}
\vspace*{0.1cm}
\caption{The control loop of a multi-strategy agent,
  responsible for the agent's daily activity. Each strategy has to
  provide an action-selection function, but other \fnsb are
  optional. See \sec \ref{sec:budgets} on how
  the agent
  changes its strategies (to find/reach goal),
  and \sec \ref{sec:fns} for the descriptions of the different \fns.}
\label{fig:loop}
\end{figure}

An (autonomous) agent is a system that senses and acts in an
environment
so that it reaches or satisfactorily maintains certain internal
states.\footnote{The internal states, such as the state of energy or
rewards, are determined based on the agent's sensing as well.}  This
is reaching food for us.
%
For a good review of the meaning of ``agents'' (and autonomy) see
Franklin et al \cite{franklin1996IsIA,Franklin1995aMinds} as well as
Wilson et al \cite{wilson1985KnowledgeGI,Strannegrd2018LearningAD}.
Figure
\ref{fig:loop} shows the basic loops for sensing and action selection
and remembering (updating, learning, ...).
We also make a distinction between an agent and a strategy. Briefly, a
strategy provides a choice of action to the agent when queried by the
agent, and given certain information such as the latest sense data as
well internal states such as memories. \sec \ref{sec:sts} describes a
range of strategies. An agent can be composite, \ie use multiple
strategies.  Even in our fairly simple task, we have found that
stand-alone pure strategies rarely work (in a plausible diversity of
environments): the agent using a purely random motion is highly
wasteful of moves, and the pure greedy agent can quickly get stuck in
dead-ends (such as corridors with no outlets). Sometimes the agent
needs to search for food (the Search task, \sec \ref{sec:tasks}), and
at other times, the food location is known, and planning a path is a
good strategy. Thus we seek 'composite' agents that use multiple
strategies.  In nature too, there is much evidence that organism use
diverse strategies (eg allocentric or map-based vs sequential
egocentric)
\cite{Wijnen2024RodentMS,Igloi2009SequentialES,Lee2024StochasticCO}.
Most our experiments use one or two strategies in an agent.

\begin{figure}[tb]
\begin{center}
\begin{minipage}{0.6\linewidth}
\hspace*{-.5cm}  \subfloat{{\includegraphics[height=2.7cm,width=9.5cm]{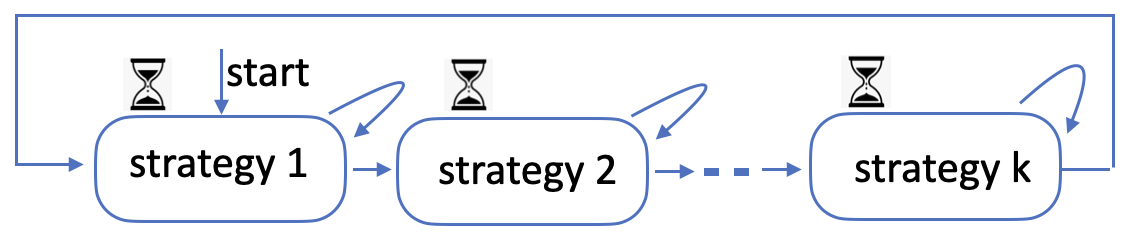} }} \\ 
   \hspace*{0.35cm} \small{ \mbox{\ \ \ \ \ (a) An agent can use several strategies in a round-robin fashion.}  }
\end{minipage}
\hspace*{.01in}\begin{minipage}{0.3\linewidth}
  \subfloat{{\includegraphics[height=2.cm,width=5.5cm]{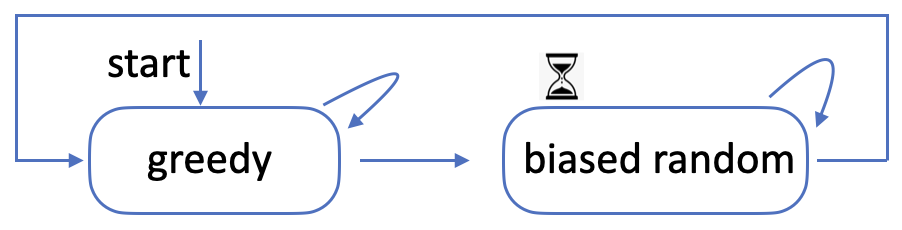} }} \\
    \hspace*{.1in}  \small{        (b) The \greedy+\biasedb composite strategy.  }
\end{minipage}
\end{center}
\vspace{.2cm}
\caption{An agent can use several strategies, or behavior modes, in a
  round-robin (fixed) ordered way in this paper (\sec \ref{sec:budgets}): (a) Each day
  the agent begins with using the first strategy. It moves to the next
  strategy (wrapping around), when current active strategy fails, or
  the strategy's time is up, until goal is reached (or all strategies
  fail). The allotted times are doubled each time it starts the list
  over in that day. (b) An example composite agent: The \greedy+\biasedb (mixed-greedy)
  agent begins with the greedy strategy and transitions to (biased)
  random in case of the failure of the greedy, and repeats this loop
  (each time, doubling the time allotted to random), until food is
  found.  }
\label{fig:round-robin} %
\end{figure}

However,
our way of using multiple strategies is not 'decision making by
committee': at each time tick, the agent takes the action selected by
exactly one designated strategy, which we call the {\bf \em active
  strategy}. The agent sticks to its active strategy until it fails,
or that strategy's current time budget is up, in which case the agent
moves to the next strategy on the (user-specified) list of strategies
(\fig \ref{fig:round-robin}).




We begin by describing how an agent is structured in more detail next,
in terms of how multiple strategies are arranged and interfaced with.
%
This structuring makes it convenient to plug in various strategies and
compare different composite agents.




\subsection{Agent's Scheduling Logic and Progressive Time Budgets}
\label{sec:budgets}
\label{sec:rr}

Our agent
uses its strategies in a {\em \bf priority} order, in a {\em \bf
  round-robin} fashion,
and under {\em \bf progressive (time) budgeting} to determine when to
change strategy.
The user specifies the ordered list of strategies to use, and the
system goes over each strategy in the fixed priority-order given, and
in a round-robin fashion. In the beginning of a day, at time tick 0,
the highest priority strategy is used: the first strategy in the list
is designated as the {\em \bf (currently) active}. The trigger to
change from one active strategy to next is either a {\em \bf times up}
or the case that the strategy {\em \bf fails}. Failure for a strategy
means that it does not return a (legal) action.  For instance, for the
case of the greedy strategy, failure means that no greedy actions is
available due to barriers. This is an indication that some other
strategy should be used. Some strategies, such as random, do not have
a failure mode.  Note that in general a strategy can become active
(reactivated) multiple times in a day (due to the round-robin
processing). If all strategies in the given list fail, the agent gives
up too. In our experiments, we always include a strategy that never
fails.

Progressive, in particular exponentially increasing, time budgets are
motivated by the consideration that the agent in general does not know
how much to stick with a given strategy (see greedy strategy, \sec
\ref{sec:greedy}).\footnote{It is possible that the agent may have
extra information, such as internal time (energy) budgets, and use
such to inform and constrain its use of different strategies.} We use a
simple dynamic budgeting technique: start the day with preset time
limits, eg 1 unit (time ticks), and double the time limit whenever the
strategy is visited again (is activated) in that day. This idea is
akin to the exponential-backoff retries when a service does not work
(temporarily) \cite{Roberts1975ALOHAPS}, \eg in distributed systems,
which helps avoid (network) congestion or overwhelming the service.
%
%
Over the days, the agent could also learn aspects such as a better
ordering of its given strategies or how much time to spent on each
strategy. Again, such a learning technique must be responsive
(adaptable) to changing environments.

In our comparisons, we create an agent, by specifying the ordered list
of one or more strategies, often two, and the initial time budget, if
any, for each strategy in the list (often either 0, meaning no budget,
or 1).



\subsection{Agent's Interfacing with the Strategies}
\label{sec:fns}

Every strategy has to implement the mandatory action selection
function, so that the agent can query it to get the action recommended
by the strategy.
In order to make a decision (\ie provide the choice of action)
a strategy may need various information,
such as the action that was last executed, even if the strategy was
not the active strategy (\sec \ref{sec:rr}), and the state of various sense
data. The agent provides this information via the following optional
{\bf \em interfacing functions} (agent$\leftrightarrow$strategy).
%
Thus a strategy may implement up to 4 other interfacing functions (in
addition to the mandatory action-selection function).
The action selection function is invoked only for the active strategy,
but these four are invoked (at appropriate times) for {\em all} the
strategies of the agent:
\begin{enumerate}
  \item The pre-action-selection function: As an example, a strategy
    could use this to record/remember what is observed around the
    agent.
  \item The post-action-selection function: useful for
    recording/remembering what action was taken.
  \item The beginning of the day function: useful for
    preparing for a new day (\eg clearing certain memories, etc).
  \item Once goal is reached, or upon-reward function: for recording
    the location of goal, etc.
\end{enumerate}

These functions are optional, and for instance the random and greedy
strategies implement none of the above. At every time tick, the agent
first invokes every strategy's pre-action selection function, if
any. After an action is selected, every strategy's post-selection
function is invoked. The agent keeps and provides common information
(such as the action taken, or the legal actions, current day, time
tick, and estimated location, ...) to all the strategies.  Because in
our experiments, we end the day once food is reached, functions 3 and
4 above could be combined. \fig \ref{tab:reqs} presents a summary of
each strategy's requirements, such as localizing and memory.




\subsubsection{Bypassing and Support for a Strategy Hierarchy}
\label{sec:bypassing}


The above interfacing functionality can support certain {\em
  hierarchical} action selection patterns.
For instance, if the food is seen next to one's cell,
the agent should just go to it, and ignore (bypass) the active
strategy's recommendation (which could be following an out-dated plan
blindly): simply invoke the pre- and post-action selection functions
of all its strategies.
%
%
%
We have observed in our experiments that this bypassing capability
lowers the number of steps to food somewhat.
In this specific scenario, the 'lower level' {\em immediacy} bypasses
the 'higher level', akin to reactive control
\cite{Thrisson1997LayeredMA,Kotseruba201840YO} (\eg avoid imminent
danger), while one can also imagine scenarios where the higher level
should over-ride (``subsume'') the lower level \cite{Brooks1986ARL}
(see section 6 in Kotseruba and Tsotsos \cite{Kotseruba201840YO}).


%
%



\section{Strategies} 
\label{sec:sts}

\begin{figure}[tb]
\begin{minipage}{0.65\linewidth}
  \begin{tabular}{ |c|c|c|c|c|c| }     \hline
     strategy & localization & smell & within day  & multi-day & planning \\ 
     $\downarrow$ &     &   & memory  & memory &  \\ \hline
    \random & -- & -- & -- & -- & -- \\ \hline
    \greedy &  -- & \checkmark & -- & -- & -- \\ \hline
    mem. \greedy & \checkmark & -- & -- & \checkmark & -- \\ \hline
    \unvisited & \checkmark & -- & \checkmark  & -- & -- \\ \hline
    \path & \checkmark & -- & \checkmark & \checkmark & -- \\ \hline
    \pmap & \checkmark & -- & \checkmark & \checkmark & \checkmark \\ \hline
    DQN & depends & -- & -- & \checkmark   & -- \\ \hline
  \end{tabular}
\end{minipage}
\hspace*{.51cm}
\begin{minipage}{0.3\linewidth}
  \subfloat{{\includegraphics[height=2.5cm,width=3.3cm]{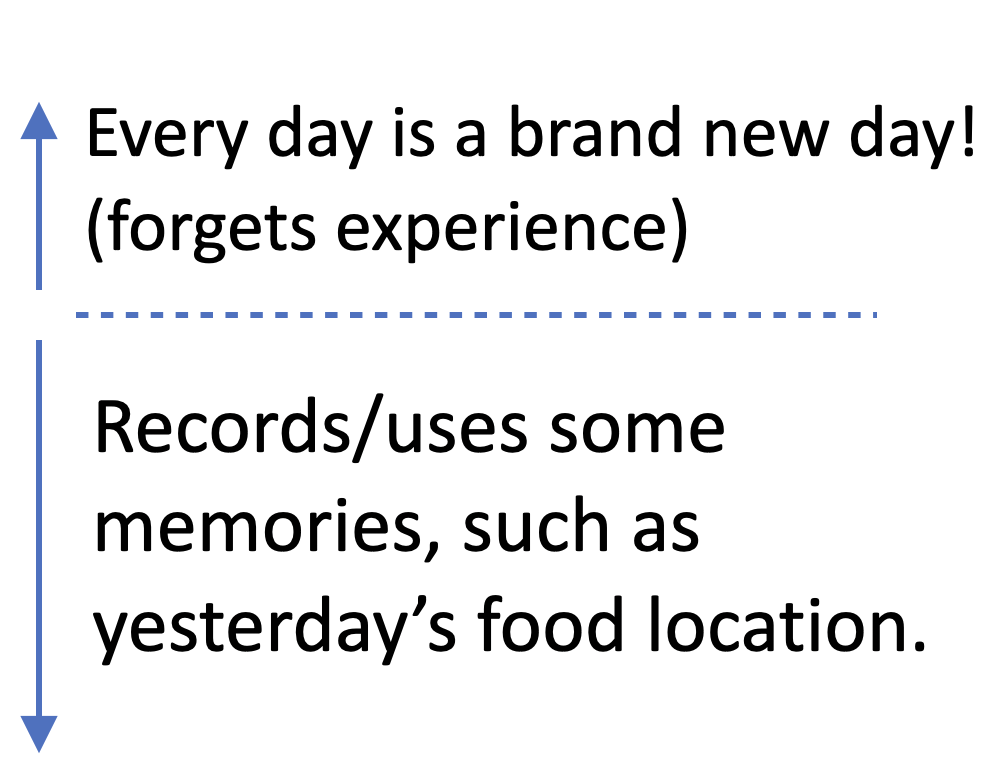} }}
\end{minipage}
\vspace{.1in}
\caption{A summary of what different strategies use or require (mainly of the
  agent, but also of the environment). \greedyb requires
  the smell ('gradient') direction. Localization, ie availability of
  the $(\tx,\ty)$ estimate of the current location for the agent, need not
  be perfect (\sec \ref{sec:sense}).  DQN's long-term memory is in its
  neural-network weights, and its input vector 
  includes current $(\tx,\ty)$ in our experiments. }
\label{tab:reqs} 
\end{figure}

A strategy provides a choice of action
when queried by the agent.
%
%
%
We next describe the strategies we experimented with, ordered
based on extent of memory use (and complexity).  See also \fig
\ref{tab:reqs} which presents a summary of each strategy's
requirements (localizing, memory, etc).

%

\subsection{Random Strategies}
\label{sec:random}



Our most basic strategy, {\bf \rand}, 
returns a legal move picked
uniformly at random (up to four possibilities). It requires no memory,
but is highly wasteful of steps. A small change, which we call {\bf
  \em \biased}, is a substantial improvement (Table
\ref{tab:steps}). \biasedb does not take a 'step back' when possible,
picking uniformly at random from the remaining legal actions (\eg if
\lftb was executed at $t-1$, \rtb is not selected at $t$, unless it's
the only legal action). This variant requires a bit of memory.
There are other variants, such as adding a further bias to move
forward when possible (in the same direction of the move at $t-1$),
but our limited experiments did not show a clear benefit.
The plain random strategy is simple and does not
%
implement any of the pre and post interfacing functions.  The \rand
strategy is a discrete random walk (the discrete version of the random
Brownian motion), while \biased and the next strategy can be viewed as
forms of self-avoiding walks
\cite{Zaburdaev2014LevyW,Humphries2010EnvironmentalCE}.



\co{
Our most basic strategy is the {\bf \em (pure) random} strategy: at
any time point, it picks a legal move (direction), uniformly at random
from the up to four possibilities, and executes that.
This strategy basically uses no memory, and is highly inefficient.

A small change, which we call {\bf \em biased random}, never takes a
step 'back', when possible: it picks uniformly at random from the
remaining legal actions. This leads to a substantial
reduction in the number of steps.  Note that biased random requires a
bit of short-term memory: do not move back to the last cell, at time
$t-1$, if possible.
}


\subsection{Greedy Strategies }
\label{sec:greedy}



The \greedyb strategy has access to the action(s) that
lead to lowering the distance to the goal (up to 2 such), and picks
one such at random. \greedyb can fail, \ie barrier(s) can block those
directions,
thus it cannot be used alone.
Like \random, \greedyb does not need any of the pre and post
functions.  A variant, {\bf memory-greedy}, does not require a smell
direction but requires the memory of the food location from the day
before. If food is fairly static, it performs similar to greedy except
when Search is required (\eg on 1st day, \sec \ref{sec:tasks}).

\subsection{Least-Visited (medium-term,  or a day's, memory)}
\label{sec:unvisited}

The {\bf \unvisited} strategy is our first strategy that makes
extensive use of what could be viewed as a type of episodic memory
(but only over a single day). 
This is also the first strategy that requires localization. This
strategy, in its prepare-for-the-day function, allocates an empty
mapping, of location to visited count (thus forgets yesterday's
information), and in its pre-action-selection function, increments the
visit-count of its current location (location estimate, $(\tx,\ty)$,
via path-integration of \sec \ref{sec:sense}).  Whenever it is the
active strategy, it picks an action that takes it to the cell with
lowest visit-count (ties broken at random).
%


Compared to random, this strategy promotes more efficient exploration,
and it is a useful strategy when (in effect) Search behavior
is required,
\eg on the first day,\footnote{Many organisms appear to have
developed so-called Levy walk and jump strategies to more efficiently
search a large expanse in finding clusters of food
\cite{Zaburdaev2014LevyW,Humphries2010EnvironmentalCE}. \unvisited in conjunction with planning
(\pmap) could be used for such search as well. }
or whenever the food location has changed, and when other strategies
(such as greedy) fail. Thus {\bf \unvisited} complements other memory
or goal oriented strategies well.







\subsection{Path-Memory (longer, over-days, memory)}
\label{sec:path}


The {\bf \pathmem} strategy remembers yesterday's path, in the form of
a mapping, from visited location (state), $(\tx,\ty)$, to
action. There can be multiple actions performed at a given location,
and the last action, selected by {\em any} active strategy, is what is
remembered: performed in the post-action-selection function, and the
previous recorded action for that location, if it exists, is
over-written.
If there is no change in the location of barriers nor food, and there
is no motion noise, the remembered mapping is guaranteed to yield a
successful path to food for the following day. This observation can be
established fairly easily.\footnote{Proof sketch: following the
remembered mapping cannot end in a cycle, \ie a node (location) being
repeated. The out-degree of every location, as key in the mapping, is
1, and the indegree, for all nodes except the start node, can be shown
to be exactly 1 as well (induction on number of hops from goal, and
arguing no node can repeat in the induction step, as the agent reached
the goal). } But note that the path may be significantly longer than
necessary.



There are a few {\bf \em execution-time} failure cases (akin to failures
in the next planning section):
\begin{enumerate}
  \item The (remembered) path, \ie the mapping, returns an illegal
    action (\eg due to a new barrier today) 
  \item The goal location is reached, but no food (\eg due to localization error)
  \item The location is not in the remembered path (from executing other strategies).
\end{enumerate}



In case of failure, the agent changes strategy. It is possible that,
for instance after exploring a bit, then landing on a later portion of
the path, following the path would be useful again. Using the
round-robin and progressive budgeting techniques allows for this
possibility in a simple way (\sec \ref{sec:budgets}).


\subsubsection{Relation to Model-Free RL}

\pathb is akin to model-free RL solutions in that the path can be
viewed as a policy, mapping locations (state features) to actions
(also, akin to sequential egocentric \cite{Igloi2009SequentialES}).
This strategy works for only one goal, and if there are multiple goals
or destinations (\sec \ref{sec:food_change}), the agent may need to
learn different paths (different mappings, or functions), which can
become both sample and space inefficient. We do not explore such
extensions here.  We also experimented with DQN \cite{mnih2013}, which
is a model-free NN approach that has been very successful in
fully-observable (Atari) game playing. We provided basic features (the
surround) and location to the network (\sec \ref{sec:dqn_exp}).

\co{
This strategy is akin to reinforcement-learning solutions in that the
path can be viewed as a policy, mapping locations (state features) to
actions.  This strategy works for only one goal, and if there are
multiple goals or destinations, the agent may need to learn different
paths (different mappings, or functions).  The next approach of
learning a (partial) map and planning is more flexible.
}



\subsection{Probabilistic Map Strategy: Remember, Learn, Plan}
\label{sec:map}



The (probabilistic) map strategy, or {\bf \pmapb} for short, records
and keeps updating memories of the barriers and food, and uses such
memories
to make (in effect) a map, that includes a start and goal, and plans a
path to the goal.
The
outcome of the planning, the computed path (plan) has a simple mapping
(hashmap) structure, same as in \sec \ref{sec:path} (a mapping from
location to action).
This strategy is the most elaborate, and the most expensive in terms
of the compute and control infrastructure it requires, but once the
memory is populated to a sufficient extent, and utilized as a map, it
is the most powerful and flexible of the strategies, since the start
and goal locations need not be fixed. The goals may not necessarily be
food, \eg exploratory unexamined destinations could be set as goals,
though we do not explore such possibilities in this paper.




\subsubsection{Making Memories}


There are two types of memories that are maintained (recorded, updated, ...):
\begin{enumerate}
\item (episodic/fast recordings) What was observed at a given time and location.
\item (slow learning) For each {\bf \em  memory-type} (described below),
  learn and maintain its predictions (distributions) and prediction
  performance (we use logloss). These predictions are selected (based
  on performance) and used when planning. 
\end{enumerate}

\begin{figure}[tb]
\begin{center}
  \begin{minipage}{0.4\linewidth}
    \hspace*{0 cm} For each adjacent location $s$: \\
    \hspace*{0.3cm}    For each of the episodic memories of $s$: \\
    \hspace*{0.6cm}       Convert the memory to a memory-type $m$. \\
    \hspace*{0.6cm}       Update performance of $m$ (logloss). \\
    \hspace*{0.6cm}       Update predictions of $m$. \\
    \hspace*{0.3cm}    Store a new episodic memory for $s$.\\ \\
    \hspace*{-0.10in} (a) Pre-action-selection for \pmapb strategy.
  \end{minipage}
  \hspace{1cm}
\begin{minipage}{0.5\linewidth}
  \subfloat{{\includegraphics[height=2.5cm,width=7.5cm]{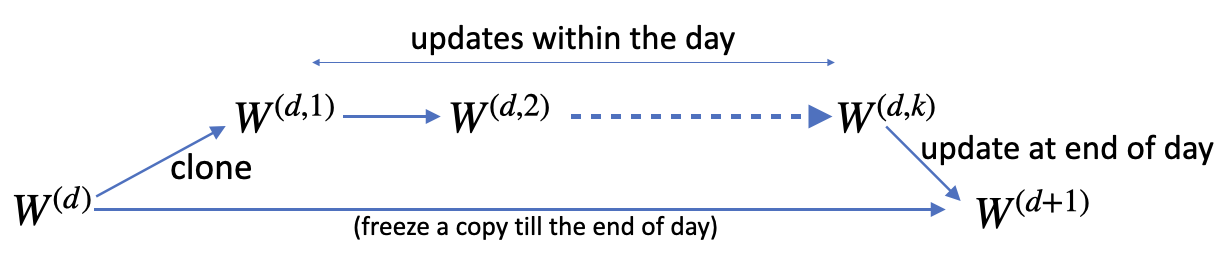} }} \\
  \hspace*{.1in} (b) two-tiered, within-day and daily, updating.
\end{minipage}
\end{center}
\vspace{.2cm}
\caption{(a) In the pre-action-selection function of the \probmapb
  strategy, for each of the 4 adjacent locations, predictions of
  memory-types are updated and a new episodic memory is formed.  (b) A
  daily estimated statistic $W^{(d)}$, such as a predicted
  distribution by a memory-type, or its loss, is cloned at the
  beginning of the day, the original does not change during the day,
  while the other, within-day clone $W^{(d, 1)}$, is used (\eg for
  planning) and may be updated many times during the day yielding
  $W^{(d, k)}$.
  At the end of the day, the within-day $W^{(d, k)}$ is incorporated
  into $W^{d)}$ to generate the next day's daily statistic,
  $W^{(d+1)}$.}
\label{fig:clone_update} %
\end{figure}

As the agent navigates the grid, at each location $(\tx,\ty)$, it
updates its episodic memory for the four locations
adjacent\footnote{The memory for the position the agent is currently
on was updated in the previous tick, or is updated on the next tick.}
$((\tx+1,\ty), (\tx-1,\ty), (\tx,\ty+1), \cdots)$: this is whether it
saw \food, \barr, or \empt \ in each of the neighboring location
(within radius 1). Memory updates are done in its pre-action-selection
function (\fig \ref{fig:clone_update}(a)).  A hashmap, from
(estimated) location $(\tx,\ty)$ (location is key) to a list of such
memory objects, is repeatedly updated.  Each such memory is a record
(object) that has a time field (day and time tick), and whether the
agent saw a \barr, \food, or \emptyb at the location. For instance the
episodic memory object <$(1,2)$, \barr, 4> means \barrb was observed on
day 4 in location $(1,2)$ (and is kept in the list of memories of
location $(1,2)$). This strategy keeps episodic memories from today,
yesterday, and so on,
kept up to some maximum number of days past, a parameter (\eg up to
5).

Such an episodic memory can have different uses for inference,
learning (long-term patterns), and ultimately prediction. For our
task, in the \pmapb strategy, the memories are used for their
predictive value when planning. They provide predictions
on where food and barriers can be. This can be viewed as an instance
of online supervised learning (different memory types predicting), and
below we explain simple ways of computing and aggregating their
predictions (for food and for barrier), although we expect much
improvement is possible, which we leave to future work.

When there exists a memory of food (\eg on day 2), this strategy
constructs a map from its memories and plans a path from home to the
remembered food location.  It helps to extract probabilities, from the
episodic memories (for barrier and food locations, described next).
There are simpler non-probabilistic approaches: keep one memory for
each location, the latest memory. The more elaborate method below
allows for more flexibility and more efficient navigation (our
experiments, in particular, when changing food locations, best reveal
the advantages of keeping probabilities).

\subsubsection{Memory-Types serve as Predictors}
\label{sec:probmem1}

Individual locations, or specific episodic memories, are not observed
sufficiently often to provide reliable predictions (the problem of
sparsity of specific, episodic, memories). Thus the \probmapb strategy
maintains
a predicted distribution for each {\em memory-type}. This is a kind of
slow statistical learning (and online and supervised\footnote{The
observation of an item, such as \foodb or \barr, provides the
supervisory feedback.}), and a memory-type allows for an appropriate
aggregation and generalization (addressing the sparsity
problem). There are a number of ways of achieving this, and we
describe one simple way, but alternatives should be explored.

In our implementation, a memory-type is specified by two components:
time to current day, or the age of the memory, and the object that was seen
(\food, \barr, and \empt).  Time to current day is the number of days
of the memory till today, so a memory-type from the current day
(today) has 0 distance (to today), while a memory from the day before
yesterday, yields a value (memory age) of 2.

Each memory-type predicts a 3-element distribution: a distribution is simply a
probability value assigned to each of the three types of objects:
\empt, \food, and \barr \ (the probabilities summing to 1.0). For
example, assume \barr \ is observed today at location $(\tx,
\ty)$. Then, two days later, this memory corresponds to the
memory-type <2, \barr>, and this memory-type provides a distribution
for the same location $(\tx, \ty)$.



In our implementation, at every time point, in the pre-action
selection function (\fig \ref{fig:clone_update}(a)),
for each of the four adjacent locations,
if there are episodic memories for that location, these memories are
converted to memory-types and the logloss from their existing
distributions (predictions) are updated, then the distributions (for
each memory-type) is updated. Note that the same episodic memory is
converted to a different memory-type on different days.


A memory could also learn and predict for other (more distant)
locations, which we expect would be useful for this task, but we do
not explore this possibility in this work.\footnote{This relates to
how an agent may 'carve' its spatial world (see \sec
\ref{sec:intro}).}. We leave exploring how memory types could be
learned and further structured to future work.





\subsubsection{Updating Predictions (Distributions)}
\label{sec:probmem2}

We perform {\bf \em two-level} updating (\fig
\ref{fig:clone_update}(b)), so that an atypical day\footnote{For
example when motion noise causes early and thus many subsequent errors
in localization.} does not have a large adverse effect on the
performance of a daily predictor for future days. In general, we want
the within day prediction to adapt fairly fast to a day's observations
and changes (in our case, localization errors),\footnote{These
localization errors (independent action noise) do not necessarily
apply to future days (independent errors). More generally, certain
patterns in one day, eg weather events, may not generalize to future.}
while the model across multiple days to be more stable.

In the beginning of the day, each {\bf daily} predictor is cloned to
create a corresponding {\bf within-day} predictor.  The within-day
predictor is used for prediction and planning in the day and is also
continually updated. At the end of the day, the predicted distribution
from the within-day predictor is used to update the daily predictor
(which initializes the next day's within-day predictor).

Each predictor is a simple fixed window moving average predictor,
where we use window size $K=10$ for a within-day predictor, and $K=5$
for a daily predictor. Each predictor keeps a queue of $K$
3-element distributions, and on an update, the oldest distribution, if
at capacity $K$, is dropped, and a new distribution is inserted. These
queuing methods are similar to those in \cite{pgs4}, where predictors
keep track of changing distributions,
with a few differences: here, we've made a closed (and small) world
assumption and, on the other hand, the input here is in general a
distribution, not a single observed item at each time. Finally, in
this task, we seek a good prediction even after one exposure (fast
learning).

The within-day predictor, corresponding to a memory-type, works as
follows. If it does not exist, it is allocated.
Upon allocation, a memory-type predicts with its corresponding object
with probability 1.0 (or near it).\footnote{More generally, the
observation may have an associated probability (from the
sensing/perception pipeline), and we could use such probabilities.}
For instance, the memory-type <2, \barr> initially predicts \barrb
with probability 1.0, or $[0, 1, 0]$, where we've taken \emptyb to
correspond to dimension 1, and \barrb and \foodb to correspond to
dimensions 2 and 3 respectively. During the day such a memory type is
updated several times. For instance, an observation of \emptyb inserts
$[1, 0, 0]$ in the queue.  Say in day 6 the agent moves to location
(5, 3) and observes \emptyb in $(5, 4)$.  The location $(5,4)$ has
memory <$(5,4)$, \barr, 5>, which has type <1, \barr>. Thus
the within-day predictor for <1, \barr> updates for the
observation $[1, 0, 0]$. Later the agent moves to $(7, 7)$, and
observes \barrb in $(6, 7)$, and $(6,7)$ also has memory <$(6,7)$,
\barr, 5> which again corresponds to memory-type <1, \barr>, and this
predictor now updates for $[0, 1, 0]$.  Say a new predictor with $K=3$
observes the sequence \mbox{[}\barr, \empty, \barr, \barr, \barr,
  ... \mbox{]}, then its predictions, after each update, would be
%
the sequence $[0, 1, 0]$, $[1/2, 1/2, 0]$, $[1/3, 2/3, 0]$, $[1/3,
  2/3, 0]$, $[0, 1, 0]$, ...
%
Imagine the barrier change rate is 0.1 (for many days). Then
memory-type <1, \barr> (a memory of a barrier from yesterday) is
expected to predict around $[0.1, 0.9, 0]$.



\subsubsection{Map Making: Sampling for Barrier Locations and Goal}
\label{sec:sample}

\pmapb uses the distributions, provided by the within-day predictors
(memory-types), to select a goal location (a destination to plan for)
and to sample barrier locations.

In each planning session, the strategy needs to have a goal
location. To set a goal, the \pmapb strategy first collects the list of all
locations that could have food with sufficiently strong probability
(above 0.01 in our experiments).  For each location, the memory
yielding the highest food probability provides the food probability
for that location.  If the list of food locations is empty (no
location has sufficient probability, eg on day 1), no planning is
possible, and the agent changes strategy (\sec
\ref{sec:budgets}). Otherwise, and in each planning iteration (see
next \sec \ref{sec:plan}), the strategy picks a location from the food
list with probability proportional to the normalized
probabilities.\footnote{Thus, if there are two locations, one with 0.5
probability, and another with 0.7, then the first gets 0.5/(0.7+0.5),
and the 2nd gets the remainder. }
Note that since we use within-day predictors (for food and barriers),
the probability of food (at a location) can change in a day, and the
list of (food) destinations can
 shrink
during the day.

We need barrier locations for planning too, and again we use sampling.
For each location, there can be multiple episodic memories (from
today, yesterday, etc). Each is converted to a memory-type. Some
memory-types are better predictors than others. We use logloss to
pick the best predictor and sample from its distribution to get
whether there is a barrier at a location.
%
If a location has no associated memory, the agent could use a prior,
the overall chance of finding a \barrb and \emptyb (which can be
updated in every day and time step). In our current implementation, we
assume empty (no barrier).  This choice may lead to more exploration
when planning.  Due to uncertainty, change and limited exploration,
this strategy, like \path, is not guaranteed to find the shortest
path, but does relatively well (\sec \ref{sec:path_exp}).\footnote{In
our current implementation, first memories are sampled, and a whole
(barrier) map is constructed, then planning is carried out (the search
for a path). An alternative implementation (possibly biologically more
plausible) would access memories as needed during planning. }

In each planning iteration, a set of barrier locations is sampled and
a goal (food location) is picked, and the strategy tries to find a
path from start (0, 0) to goal, further described next.

{\bf Brief Discussion of Updating and Sampling.} Food is sparse (in
the environments of our experiments) and on the other hand important
to the agent, and perhaps that explains the way we sample barriers
(for planning) and food location (for setting a goal) are different:
for barriers, we pick a best predictor, while for food, we use the
highest probability prediction. It would be good to further clarify
these underlying differences in addition to improving the updating
(learning) and sampling techniques.


\subsubsection{Planning a Path Given a Map (Constructed from Memory)}
\label{sec:plan}


%
Whenever the \pmapb strategy is activated (such as the beginning of the day)
and whenever there is a problem in plan
execution (such as an unexpected barrier),
%
 the strategy performs planning, which, in case of success, yields a
 (new) path or plan to goal (same structure as the path agent). The
 strategy executes according to that plan/path until goal is reached,
 or until the next (execution-time) failure (described below) or times
 up. The planning algorithm is simple: pick and set a goal destination
 and extract (sample) a barrier map (\sec \ref{sec:sample}), and use a
 search algorithm. We use $A^*$ best-first search \cite{Hart1968AFB,russell2021artificial}.


There are two main types of failure:
\begin{enumerate}
\item Planning time failure, \ie no food (no destination to go to, \eg
  on day 1) or (seemingly) no path to food (within the time budget). The
  agent needs to move on to another strategy (\sec \ref{sec:budgets}).
\item Execution time failure, such as new barriers (see
  \ref{sec:path}). The strategy replans in this case (unlike the
  \pathmem strategy).
\end{enumerate}

When planning, the strategy repeats path-finding ($A^*$ search) a fixed
number of iterations (5 in our experiments). As soon as one iteration
yields a path, that path (plan) is returned.  In each path-finding
iteration, a goal location and barriers are sampled anew. A
path-finding search aborts (stops that iteration) if the current path
estimate exceeds the budget on number of steps (given to the
strategy). Since barriers are sampled at the beginning of a planning iteration,
a subsequent path-finding iteration may find a path (while previous
ones failed).

\subsection{The Oracle Strategy}

The \oracle strategy is meant to provide a reference point, the best
(lowest) number of steps on a given day and environment.  The strategy
always has the complete up-to-date map (of barriers and food), as well
as the true location of the agent, and plans accordingly (similar to
\probmap, uses $A^*$ for planning, see \sec \ref{sec:plan}). Its only
limitation is that when motion-noise is present, the strategy may need
to replan when the agent ends up in a location not in its current
plan.

\co{
\subsection{Agents we experiment with}


The agents we focus on:
\begin{enumerate}
  \item mixed-greedy: combines the greedy strategy with biased random
  \item unvisited
  \item path-memory$^+$: combines path-memory with unvisited
  \item prob. map : combines \probmap with unvisited
\end{enumerate}

In some experiments we also report the optimal cost (of a best path).

}





\co{
\subsection{RL Agents}


Model-free. deep RL ,  DQN agents \cite{mnih2013}.

Other: transformer based DTQN, etc.
 
And discussion of the MERLIN architecture and agent \cite{merlin2018}
}


\subsection{A Discussion of Types of Memory} There are many
ways to classify memories, such as episodic, declarative, semantic,
associative, working, short-term \vs long-term, internal \vs external
(\eg humans using notebooks as external memories), biographical, and
so on. We noted that \biasedb uses a little memory, and control
strategies such as round-robin and progressive need some memory for
their operation. One aspect that distinguishes these from the memory
used by \unvisited, \path, and \pmapb is that the latter's episodic
memory requirement grows in general with experience or the history of
the agent (with the spatial expanse explored and/or over time), while
the former (often control memory) is fixed (constant or near
constant).

\section{Experiments}
\label{sec:exps}


\begin{table}[t] \center
  \begin{tabular}{ |c|c|c|c|c|c| }     \hline
    & mean-mean & med-mean & med-med & max-mean & max-max    \\ \hline
    \random  &  1965  $\pm 617$ & 1815 & 1200 & 3.5k  & 30K  \\ \hline
 \biasedb (\random) & 536 $\pm 163$ & 479 & 329 & 1056 & 5.7k  \\ \hline
 \greedy+\biasedb (mixed-greedy) & 200 $\pm 110$  & 167 & 37 & 541 & 3.3k  \\ \hline 
 \unvisited  & 250 $\pm 49$  &  252 & 198 & 395  & 1.5k   \\ \hline
 \greedy+\unvisited  & 99 $\pm 44$  & 89  & 33 & 231  & 1.8k   \\ \hline 
 \path+\unvisited  & 209 $\pm 48$  & 203 & 155 & 325  & 1.7k   \\ \hline 
 {\bf \probmap+\unvisited }  & 80 $\pm 35$  & 79 & 35 & 230 &  1.7k \\ \hline 
 \oracle (theoretical minimum)   & 15.5 $\pm 1.1$  & 15.2  & 14 & 20.2 & 51  \\ \hline
  \end{tabular}
  \vspace*{.2cm}
  \caption{Performance, \ie the number of steps to goal (lower is better),
    of a range of strategies, from little or no memory, to extensive
    memory and computing (planning). The mixed \pmapb overall does the best
    (\oracle gives the minimum in the impractical case of complete
    knowledge).
    Experimental settings: 15x15 grids, 50 environments, 20 days each,
    0.3 barrier-proportion, 0.1 change-rate, and 0.02 motion
    noise. Mean-mean is average number of steps over the 20 days and
    then average over the 50 initial environments, while med-mean is
    the mean of the (50) medians. }
  \label{tab:baseline_comparisons}
  \label{tab:steps}
\end{table}
 
We used PYGAME (\mbox{www.pygame.org}) to develop the environments and
to visualize (code available on GitHub \cite{barriers_code}). Our experiments
involve 3 nested loops.  With an outer loop of $k_1$ trials (\eg 50),
we generate initial environments (grids with certain barrier
proportion).\footnote{
Beyond a barrier proportion of 0.3 often no path to food exists on 15x15 (under
random generation), and so our experiments go up to the 0.3 level. }
In an inner loop of $k_2$ iterations we generate
days: the initial environment is changed somewhat day after day (with
a positive change-rate). Finally, within a day, the experiment
continues for $k_3$ steps until the agent, starting from home, reaches
food: $k_3$ depends on agent efficacy.  We average $k_3$ over days and
then over the environments, but also report daily medians
(median-means: average the median over the environments) and maximums
too (max-means), see Table \ref{tab:steps}. For instance, max-max
refers to taking the maximum of the ($k_3$) steps over all the days
and environments (the worst day). Each row of the table took seconds
to complete on a Mac laptop.

\co{ The agent begins at the center of the grid, and the food is at
  lower right.  We see that as we add memory, the performances,
  specially the worst cases, very bad days (max-max), improve
  substantially.  This is on 15x15 grids, thus a good path should be
  about 15 steps long, modulo motion noise and barriers, and the
  \oracle's performance, 15.5 steps, reflects that (Table
  \ref{tab:steps}).

For composite agents, we set no budget on a strategy that can fails
(\eg \greedy and \pmap), so the agent sticks to those strategies until
they fail. And we put an initial budget of 1 on the other exploratory
strategy (\biasedb and \unvisited), so within a day their budget
doubles each time they are revisited during round-robin activation.
}

%
For the composite agents, we used initial budgets of 5 steps on all
strategies, doubling each time a strategy is reactivated (\sec
\ref{sec:rr}).\footnote{In the shorter version of the paper
\cite{flatland2}, we used '0,1', \ie no budget on \greedy, \path, and
\pmapb (and initial budget of 1 for \biasedb and \unvisitedn), which
yielded somewhat fewer steps overall. However, \pmapb used a default
initial budget of 50 in that case.  To make the results more grid-size
(distance to food) agnostic, we use initial 5 for all strategies.  }
\greedy+\biasedb (or {\em \bf mixed-greedy}) means start with \greedyb then
\biasedb in the round-robin fashion.
The agent begins at the center of the grid, and the food is at the
same corner on all the days, such as the lower right
(except for \sec \ref{sec:food_change}).

We begin this section by discussing the results in Table
\ref{tab:steps} next. We then report results when changing parameters
such as grid size (distance to food) and motion noise (comparing
\greedyb to \pmap).
%
We include further comparisons with the \pathb strategy as well as a
model-free RL technique. Finally we look at performance when the goal
location is changed following a few patterns, and conclude with
reporting on several statistics including the size of (episodic)
memory consumed and the number of planning invocations (in a few
settings).

\subsection{Memory Helps!}

\co{ Memory helps!  \biasedb (with a bit of memory, \sec
  \ref{sec:random}) does substantially better than \random. The best
  of memory strategies substantially beats \greedy+\biased, the bast
  of sensing minimal-memory strategies, and in particular we get
  increased robustness (the extent of 'bad' days decrease). If we
  ignore the first day or so (the Search problems), this gap grows
  (mean-mean of \probmapb improves to 50).

The best combination (of smell-direction and memory),
\probmap+\greedy+\unvisited gets a mean of 40 (not shown in the
table).
}

Table \ref{tab:steps} reports results on 15x15 grids, thus a good path
should be close to 15 steps long, modulo motion noise and barriers, and
the \oracle's performance, 15.5 mean steps (the minimum possible), reflects
that.
We can see that as we go down the table (use strategies that roughly
use increasing more memory), the performances, specially the worst
cases, on very bad days (max-max), improve substantially. \biased,
which is \randb but with a bit of memory
that prevents the agent from reversing (\sec \ref{sec:random}), does
substantially better than \random.

The best of the memory-based strategies substantially beats
\greedy+\biased, and in particular we get increased robustness. If we
ignore the first day or so (the Search days), this gap grows
(mean-mean of \probmapb improves to 50). Note that the expected
performance of \greedy+\biasedb can not change from day to day (no
remembering). The best combination (of smell-direction capability and
memory), \ie
\probmap+\greedy+\unvisited, gets a mean of 56 (not shown in the
table).



\subsection{Interaction of Size, Change-Rate, and Motion Noise}
\begin{figure}[!htbp]
\begin{center}
\begin{minipage}{0.4\linewidth}
  \subfloat{{\includegraphics[height=5cm,width=5cm]{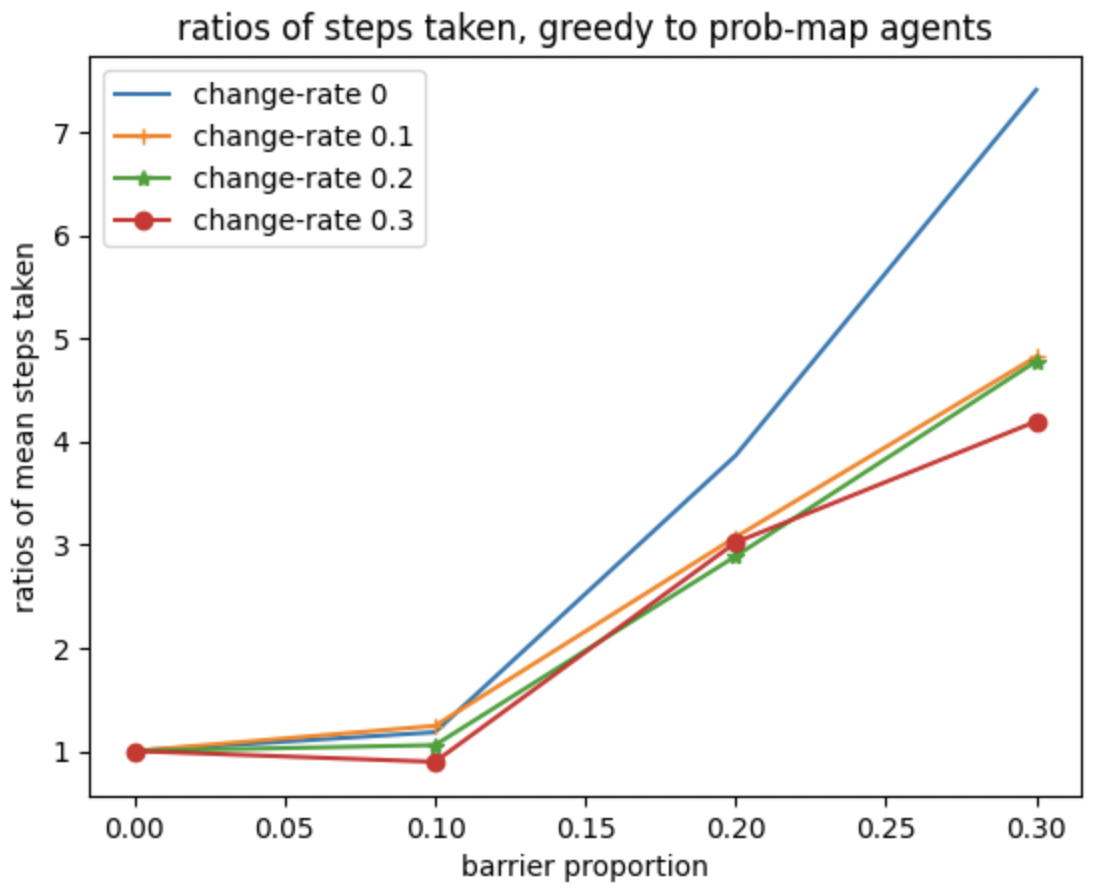}}  }
\end{minipage}
\hspace*{.11in}\begin{minipage}{0.4\linewidth}
  \subfloat{{\includegraphics[height=5cm,width=5cm]{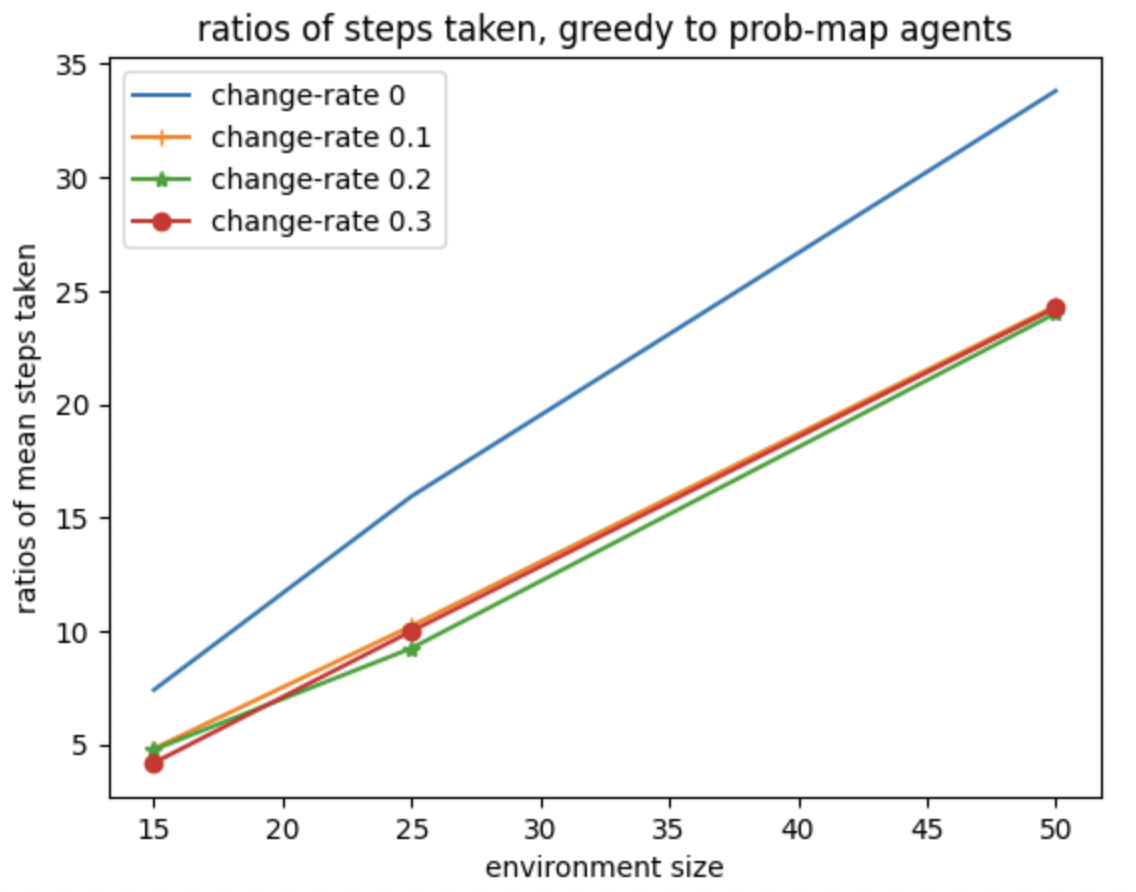} }}
\end{minipage}
\end{center}
\vspace*{-.2cm}
\caption{Efficiency gains, \probmap+\unvisited compared to
  \greedy+\biased, mean steps over 20 days and 50 environments.  The
  ratio of steps taken, \greedy+\biasedb to \probmap+\unvisitedn, (a)
  as we increase barrier-proportion ratio, for different change rates
  (on 15x15, 0 motion-noise), and (b) as we increase environment size
  (from 15x15 to 50x50), keeping barrier proportion at 0.3. }
\label{fig:ratios} %
\end{figure}

\begin{figure}[tb]
\begin{center}
\begin{minipage}{0.4\linewidth}
\hspace*{-1.4cm}  \subfloat{{\includegraphics[height=5cm,width=6cm]{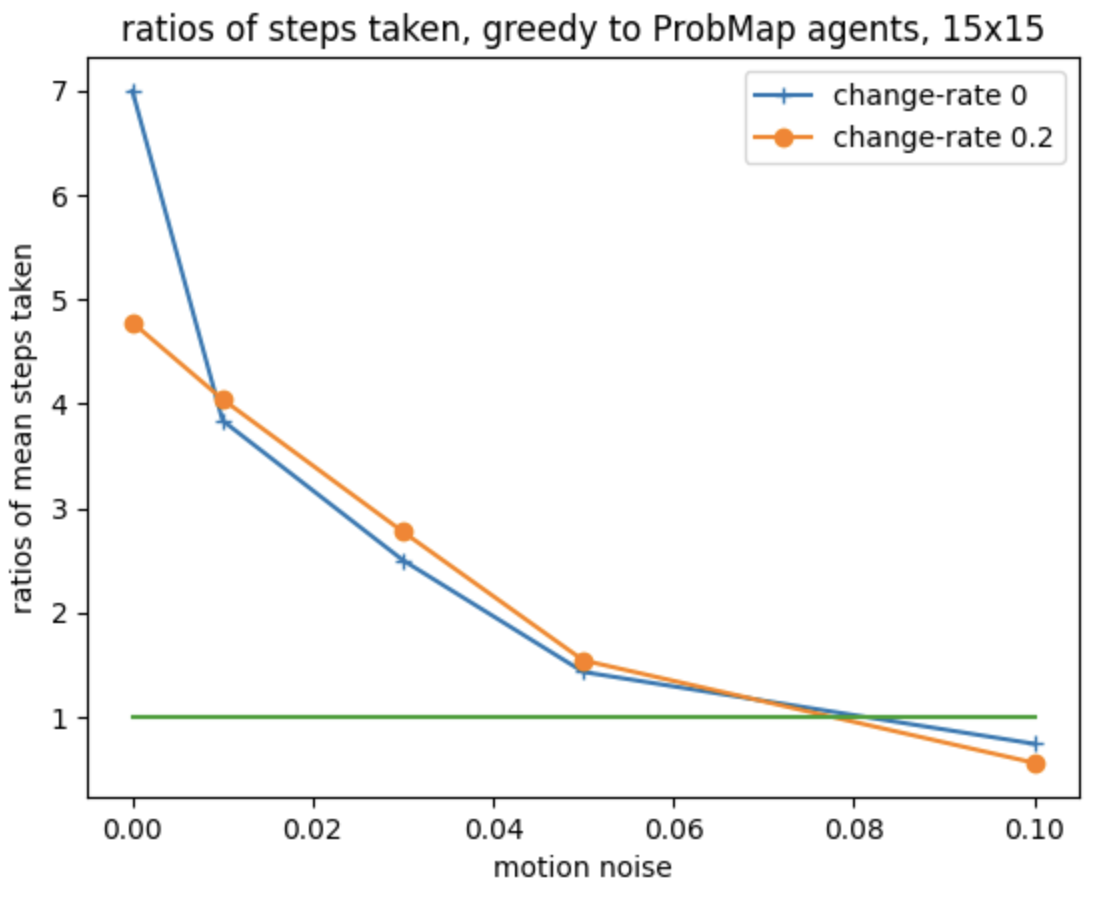}}  }
\end{minipage}
\hspace*{.11in}
\begin{minipage}{0.4\linewidth}
  \subfloat{{\includegraphics[height=5cm,width=6cm]{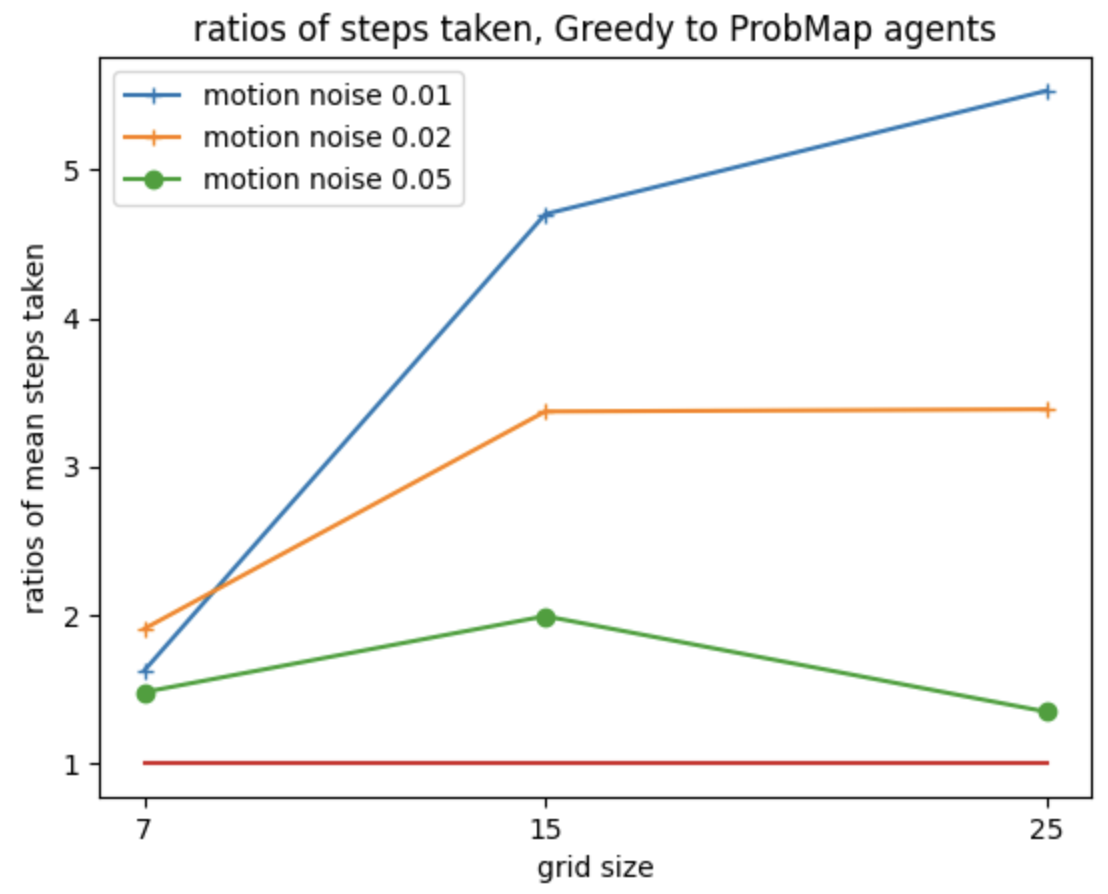} }}
\end{minipage}
\end{center}
\vspace{-.2cm}
\caption{As motion-noise is increased, the advantage of \pmap+\unvisited
  compared to \greedy+\biasedb degrades. Left: on a fixed 15x15 grid (barrier
  proportion 0.3), Right: as we increase the grid size (distance to
  goal). However, with 0 motion-noise, the performance advantage grows
  with distance (as in \fig \ref{fig:ratios}). }
\label{fig:ratios_motion_noise} %
\end{figure}

 \fig \ref{fig:ratios} shows that when grid size or barrier portion
 difficulties are raised, with low motion-noise, the relative gain of
 \pmapb compared to \greedyb grows, to over 30x (substantial gains, as
 distance grows, under no noise). \fig \ref{fig:ratios_motion_noise}
 shows, on the other hand, that motion noise has the reverse effect,
 and at some point, \greedyb can outperform. Increasing the grid-size
 can compound this.


\subsection{Keep Updating (Continual Remembering, Learning, ...)}
\label{sec:continual}

\begin{figure}[tb]
\begin{center}
\begin{minipage}{0.4\linewidth}
\hspace*{-1.5cm}  \subfloat{{\includegraphics[height=5cm,width=5.5cm]{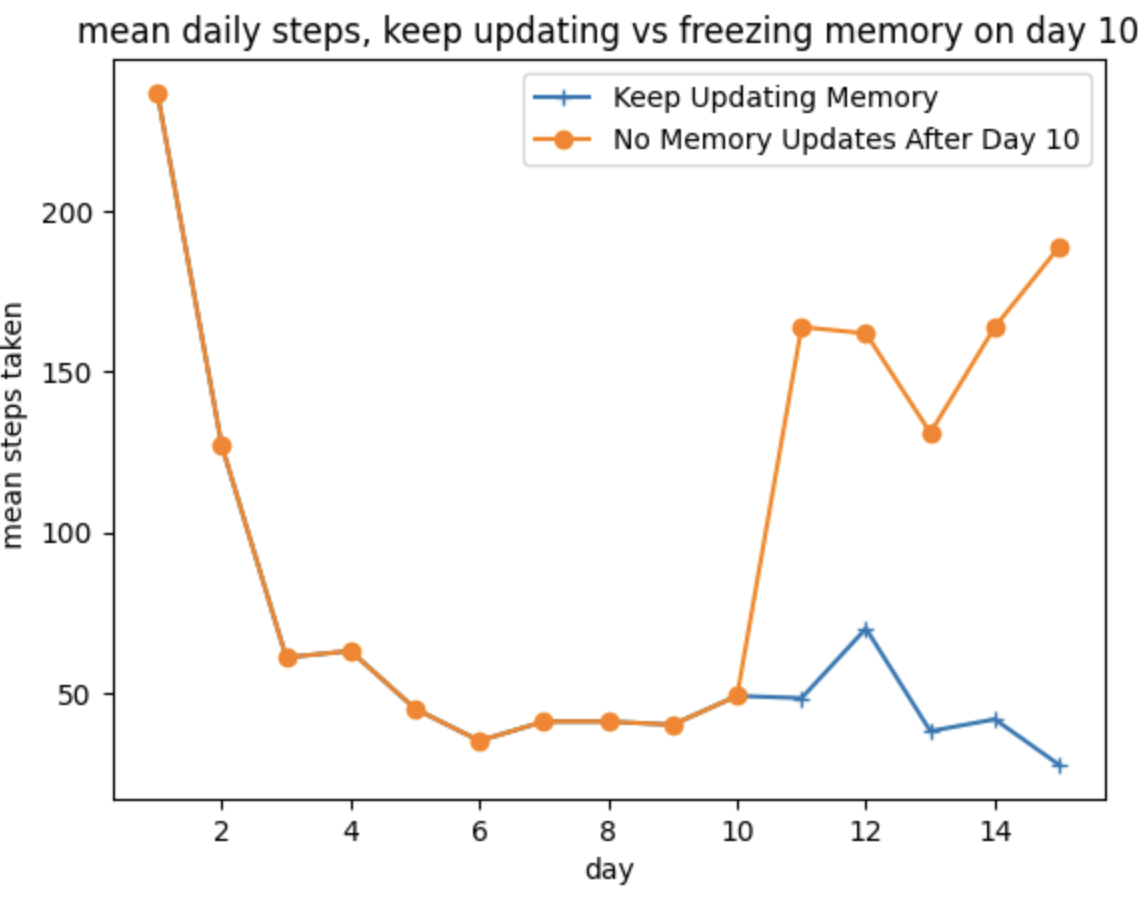}}  }
\end{minipage}
\begin{minipage}{0.4\linewidth}
  \subfloat{{\includegraphics[height=5cm,width=5.5cm]{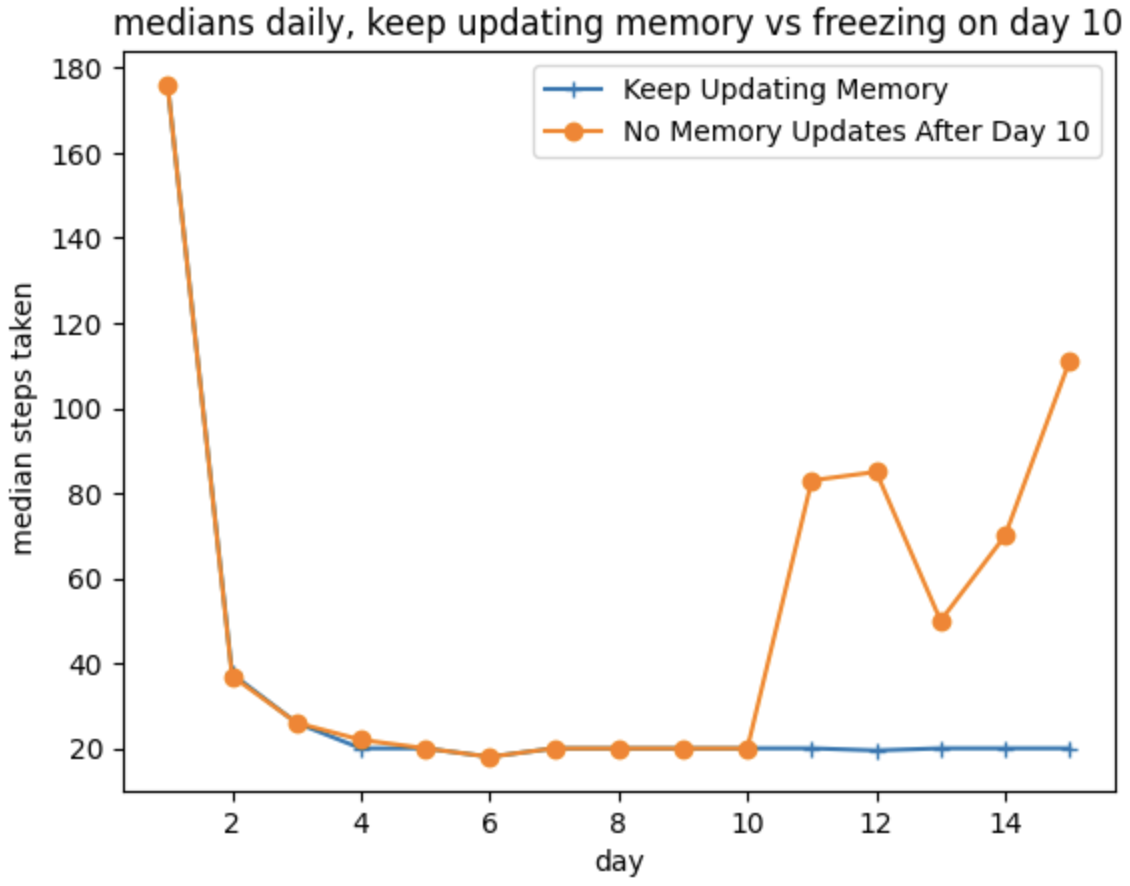} }}
\end{minipage}
\end{center}
\vspace*{-.2cm}
\caption{Continual learning (memory updates) vs freezing memory (\ie
  stopping updates after day 10). Same setting for Table
  \ref{tab:steps} except for 150 environments. Daily steps, means (left) and
  medians (right), as a function of day (each point is the mean or
  median, 150 values, of number of steps taken to food on that day).}
\label{fig:daily_steps} %
\end{figure}

\fig \ref{fig:daily_steps} shows that even though the performance of
\pmap+\unvisited may appear to plateau, the agent needs to keep
remembering and learning under (barrier) change to preserve its
performance (\eg similar findings on the need for continued updates in
\cite{pgs4}). Note that the performance of the strategies that do not
use episodic memories (\ie \rand, \greedy, and \unvisited) does not
change (improve) over days.


\subsection{Further Comparisons with the \pathb Strategy} 
\label{sec:path_exp}

The \pathb variant trails the \greedyb variants in the setting of
Table \ref{tab:steps}. If we lower the motion noise and change rates
to 0, \pathb gets a mean-mean of 39 beating 70 for \greedy+\unvisited
(and \probmapb gets 18).

\pmapb is more robust to change compared to \path. Neither approach is
guaranteed to find a shortest path, but as Table
\ref{tab:path_compare} shows, under 0 change and motion noise,
\probmap+\unvisited tends to find significantly shorter
paths. Interestingly, as we increase the barrier proportion, the
relative under performance of the \pathb variant
(\probmap+\unvisitedn) first shoots up, then as more barriers are
added goes down again.

\begin{table}[t] \center
  \begin{tabular}{ |c|c|c|c|c||c|c|c|c| }     \hline
   mean \& max steps & \multicolumn{4}{|c||}{on 15x15} & \multicolumn{4}{c|}{on 30x30} \\ \hline
   barrier prop. $\rightarrow$ & 0.0 & 0.1 & 0.2 & 0.3  & 0.0 & 0.1 & 0.2 & 0.3    \\ \hline
    \path+\unvisited & 28, 28  & 44.6, 110  & 35.2, 68 & 28.7, 58 & 62, 62 & 91.4, 206 & 92, 254  & 67.3, 172   \\ \hline
  \probmap+\unvisited & 14, 14  & 14.5, 24  & 15.8, 46 & 16.4, 52 & 28, 28  & 29.2, 52 & 31.7, 66 & 36.8, 130 \\ \hline
   \oracle &  14, 14 & 14.1, 16 & 14.6, 20 & 16.0, 36 & 28, 28 & 28.0, 30 & 28.5, 36  & 32.3, 76  \\ \hline
  \end{tabular}
  \vspace*{.2cm}
  \caption{The average and maximum (max-max) number of steps to goal
    on 15x15 (left) and 30x30 (right), on 50 environments, average and
    max taken over the last 10 days of 30 days in each environment, as
    barrier proportion is increased, with 0 change-rate and 0 motion
    noise. \probmap+\unvisited remains much closer to the performance
    of the \oraclen. Interestingly, the performance of \pathb is not
    monotonic with increasing barrier proportion (unlike the other two
    strategies): it first goes up (degrades) then comes down
    (improves). }
  \label{tab:path_compare}
\end{table}

\subsubsection{A Model-Free NN-Based Strategy} 
\label{sec:dqn_exp}

We also experimented with DQN \cite{mnih2013}, a model-free RL
technique which performs very well on a range of (fully-observable)
Atari games. The agent is given the location, and the surrounding
barrier and food information (radius 1, as other agents). On 15x15
grids, and no barriers nor uncertainty (\ie no change in barriers and
no motion noise), the agent learns a good path eventually: a learning
rate of 0.002 works best (compared to 0.01 and 0.001),\footnote{Other
parameters: batch size of 5, gamma of 0.9, eps\_start of 0.9, end of
0.05 with decay of 10. We also experimented with changing those
parameters. } and yields a mean of 80 (median 40) for 20 days, and
mean of 30 after 50 days, and 20 for 200 days.
%
However, as we increase barrier proportion, \eg to 0.1, the learning
becomes unstable, and on some environments the number of steps even
after several days is in the 10s of thousands, occurring frequent
enough that we could not often run such on 20 environments. If we set
the change rate to 0.1, the fraction of environments when the agent
hangs (has many days with say over 50k steps) goes up. The DQN agent
needs both to find the reward, and to pass that information gradually
to other cells nearby, and implicitly learn that barriers block. All
this, reward propagation, takes times (many steps and days), and if
there is change (in goal or barriers), issues of learning stability
arise.

\co{
\subsection{Interaction of Size, Change-Rate, and Motion Noise}

\fig \ref{fig:ratios} shows
that when grid size or barrier portion difficulties are raised, with
low motion-noise, the relative gain of \pmapb compared to \greedyb
grows, while \fig \ref{fig:ratios_motion_noise} shows that motion
noise has the reverse effect, and at some point, \greedyb can
outperform, and increasing the grid-size can compound this.
}

\co{
\subsection{Performance of Path} 

\pathb trails the \greedyb variants in the setting of Table
\ref{tab:step}. If we lower the motion noise and change rates to 0,
\pathb gets a mean-mean of 39 beating 70 for \greedy+\unvisited (and
\probmapb gets 18). \pmapb is more robust to change compared to \path.
}

\subsection{Changes in Food Location}
\label{sec:food_change}


Recall that the episodic memories recorded, of barriers and food, have
a time field too (the day and tick is recorded), which can be helpful
for discovering when there is a pattern to the changes in food or
barrier locations.  In this section, we explore how the \pmapb
strategy, using such memories, handles certain changes, either random
or systematic, to food location, from day to day.
In these experiments, the food is placed on one of a fixed list of $k$
corner locations: $k \in [2,3,4]$ and the list is not changed for the
duration of the experiment. In one version of the experiments,
``uniform'',
the food location is picked uniformly at random from the list. As
always, the agent is not given any information on food location.
Here, to perform well, a memory-based agent needs to remember likely
places where food could be.  In another set of experiments,
``round-robin'', we change the food among the $k$ in a strict
round-robin manner, thus when $k=2$, the food oscillates from one
corner (odd days) to the other (even days). In the round-robin cases,
the predictiveness of an agent's food memory locations will not be
monotonic with time: when $k=2$, the memory from two days ago is
better than the food memory from yesterday (recall that each
memory-type only provides, or predicts, a distribution for its own
location in our current techniques, \sec \ref{sec:probmem1}).

Table \ref{tab:goal_change} summarizes the performance of
\pmap+\unvisited, when we change the memory capacity of \pmapb from
the default of 5 days (on 15x15 grids with barrier portion of 0.3).
We observe that keeping only one day underperforms drastically in all
cases, while keeping 2 improves a bit, in particular for the
oscillating $k=2$ case (where exactly two memories from past 2 days is
sufficient), but insufficient for the uniform $k\ge 2$.  We note that,
in general, the uniform case is harder than the oscillating case (in
these experiments), as older memories may be needed to best keep
track of the possibilities of where food may be. If the agent keeps
only a few days' worth of memory (if it doesn't stretch long enough
into past), a run of several consecutive days where the food is always
in one corner completely erases the possibility (knowledge) that the
food could be on another corner as well, and when the agent doesn't
find the food in the expected corner, the problem becomes a Search
problem.

We note that, as in \sec \ref{sec:continual}, the agent is indeed
learning. For instance, for the case of $k=2$ round-robin (oscillate),
if we average the number of steps on the first 10 days, we get (mean)
over 80 steps, and if over first 5 days, we get 130 steps. Finally, to
compare to Table \sec \ref{tab:steps}, if we also introduce 0.02 noise
in motion, the mean steps goes up to around 100 (averaged over last 10
days). Finally, as expected, our \pathb strategy (designed for a
single goal) does not do well in these multiple goal experiments,
getting a mean of over 200 steps (average over last 10).


\begin{table}[t] \center
  \begin{tabular}{ |c|c|c|c|c| }     \hline
  $\downarrow$ memory kept & $k=2$, round-robin & $k=2$, uniform &
    $k=4$, round-robin & $k=4$, uniform \\ \hline 1 day & 178 & 124 &
    182 & 179 \\ \hline 2 days & 28.4 & 81.1 & 192 & 165 \\ \hline 5
    (default) & 34.7 & 59.2 & 33.1 & 159 \\ \hline 10 days & 29.7 &
    55.3 & 41.4 & 129 \\ \hline
  \end{tabular}
  \vspace*{.2cm}
  \caption{The performance (mean-mean) of \pmap+\unvisited as food
    location is changed  from day to day (\sec \ref{sec:food_change}):
    k=2 means over 2 corner locations, and k=4 is over 4 (15x15 grids,
    change-rate of 0.1, barrier-portion of 0.3, motion noise of 0, 50
    environments, 30 days) and averages are over the last 10 days (of
    30 days).  Oracle and \greedy+\biased, both being insensitive to
    the food location pattern, get respectively around 15.5 and 160s
    in this setting.  Keeping episodic memories and memory-types helps
    the agent to discover the food pattern lowering its step
    counts. When more memory is kept, the agent can handle larger
    $k$. }
  \label{tab:goal_change}
\end{table}

\subsection{Brief Exploration of a few other Statistics}

We go back to the setting of Table \ref{tab:steps} (unless specified),
in particular one corner food location, and focusing on the
\pmap+\unvisited strategy, and report statistics on a few other
aspects of that strategy.

\subsubsection{Progressive Budgets}
\label{sec:exp_progr}

Progressive budgeting doubles the time budget of a strategy (if not
unlimited) each time the strategy is activated in a day (\sec
\ref{sec:budgets}). This is a convenient way of flexibly increasing
the time allocation without knowing how far the food is or the level
of the (barrier) complexity of the environment.  As Table
\ref{tab:progressive} shows, setting the budget to a fixed value that
is too small or too large can lead to too many steps: with fixed
budget of only 5 on both \greedyb and \biased, the corresponding
greedy agent gets stuck (switching to greedy too early leads to
heading back to a dead end again and again). The table's results are
on 15x15 (same setting of Table \ref{tab:steps}. On 30x30 (not shown),
progressive budgeting's mean performance for the greedy agent is
often double that of well-chosen fixed-budgets, while for the \pmapb
agent, the performances remain comparable.  Thus there is likely room
for significant improvement: for instance, learning good beginning of
day budgets, from what levels worked on previous days, as well as
perhaps tuning the multiplier (rather than the default of 2), may
improve the results.  We also note that at times there can be a trade
off between mean \vs median performance (in number of steps): while
the median number of steps may go down (\eg when lowering the budget
of exploration), the average (and maximum steps over the days) may go
up.

\begin{table}[t] \center
  \begin{tabular}{ |c|c|c|c|c|c|c|c| }     \hline
 budgets $\rightarrow$   & progressive (5,5) & 5 & 10 & 20 & 40 & 60 & 80 \\ \hline 
 \greedy+\biased  & 200 & -- & 263 & 127 & 164 & 209 & 229 \\ \hline
 \pmap+\unvisited & 80 & 89 & 92 & 83 & 94 & 118  & 117 \\ \hline
  \end{tabular}
  \vspace*{.2cm}
  \caption{The default ('5,5') progressive budgeting (first column)
    \vs different fixed time budgets (\eg the 2nd column is fixed
    budget of 5 on both \greedyb and \biased). The setting is as in
    Table \ref{tab:steps}, and mean-means are reported. Progressive
    budgeting allows for a robust way of using multiple
    strategies. \greedy+\biasedb does not finish with a fixed budget
    of 5 (\sec \ref{sec:exp_progr}). }
\label{tab:progressive}
\end{table}

\subsubsection{Memory Consumption and Replannings}
\label{sec:exp_mems}

\begin{table}[t] \center
  \begin{tabular}{ |c|c|c|c|c| }     \hline
    change-rate and motion-noise $\rightarrow$
    & 0, 0 & 0.1, 0 & 0, 0.02 & 0.1, 0.02  \\ \hline 
    \pmap+\unvisited 15x15 & 1, 309, (90, 20) & 3, 357 (90, 31)
    & 3, 433 (100, 59)  & 5.5, 517 (101, 58) \\ \hline
 \pmap+\unvisited 25x25 & 1, 545, (274, 41) &  5, 722, (274, 86)
 & 7, 1.7k, (376, 241) & 10, 1.7k, (365, 267) \\ \hline
  \end{tabular}
  \vspace*{.2cm}
  \caption{Under a few change-rate and motion noise levels, the median
    number of planning invocations (computed over days 2 onwards), the
    mean number of episodic memories at day 20 (middle number), and
    the number of cells in the visit counts on days 1 and 2, are shown
    (environment parameters as in Table \ref{tab:steps} unless
    specified). As the task gets harder, more memory is required and
    additional replannings are carried out (\sec
    \ref{sec:exp_mems}). The visit-count goes down significantly on
    day 2, specially with low motion noise (with a more focused
    path). }
\label{tab:exp_plans_mem}
\end{table}


When \pmap+\unvisited is used, planning is invoked
at the beginning of the day (except for day 1) and every time the
current plan fails (\eg a new barrier) and whenever the strategy is
reactivated in a day. We expect the number of plannings to increase as
change rate or motion noise are raised.
Table \ref{tab:exp_plans_mem} shows
this for 15x15 and 25x25, confirming our expectations. The table also
shows memory consumption: the number of episodic memories, on day 20
(averaged over 50 environments), as well as the visit counts, on days
1 and 2.\footnote{All these statistics are reported in the end or
start of the next day function (\eg before the visit counts are reset
for next day).} Since we keep up to 5 days-past worth of memory, and
each memory corresponds to one location, the maximum episodic-memory
consumption is up to $6\times 25\times 25 \approx 4k$.

We observe that increasing motion noise significantly increases memory
consumptions as well, and the visit counts can be significantly lower
on day 2 and subsequent days (following a path), compared to day 1
(search for food).

\co{
\begin{table}[t] \center
  \begin{tabular}{ |c|c|c|c|c| }     \hline
    change-rate and motion-noise $\rightarrow$
    & 0.1, 0 & 0, 0.02 & 0.3, 0.0 & 0.3, 0.02  \\ \hline 
 \pmap+\unvisited 15x15 & 3 & 3 & 5 & 6 \\ \hline
 \pmap+\unvisited 25x25 & 4.5 & 6 & 7  & 12 \\ \hline
  \end{tabular}
  \vspace*{.2cm}
  \caption{Day 0 has no planning. Under a few positive change-rate or
    motion noise levels, the table shows the median number of
    plannings over subsequent days (with 0 change and motion noise,
    exactly one planning occurs). }
\label{tab:replans}
\end{table}
}

%

\subsubsection{State of Memories (Remembered Food, etc)}
\label{sec:exps_food_stats}

With 0 motion noise, the number of remembered food locations stays at
1 on days 2 onwards, irrespective of (barrier) change-rate or
grid-size, as expected. With motion noise of 0.01, the remembered food
locations goes to 2 to 4 on subsequent days on 15x15 (on day 3 it can
be up to two locations with positive probability, etc), and with
motion noise of 0.02, we observed 5 or 6 (\eg on day 10). When we
increase grid-size to 25x25 (distance to food), there is more
opportunity for error in localization, and at motion noise levels 0.01
and 0.02, we get many more 5s and 6s locations than for 15x15 (up to 5
past days memories by default).  An example evolution of food
locations with the probabilities day by day is shown in Table
\ref{tab:goals}.  Note that a memory may not necessarily get updated
during a day (when not observed), specially that \foodb is sparse, and
we keep up to 5 days memory by default. As food is important and
sparse, it may be good to have a differential policy as to which
memories are kept longer.


\begin{table}[t] \center
\begin{tabular}{|l| } \hline
 day 2: ((13, 16),  1.00) \\
 day 3: ((14, 15), 1.00), \ ((13, 16), 1.00) \\
 day 4: ((14, 15), 1.00), \ ((13, 16), 1.00), \ ((14, 14), 1.00) \\
 day 5: ((14, 15), 1.00), \  ((13, 16), 1.00), \  ((14, 14), 1.00) \\
 day 6: ((14, 15), 1.00), \  ((13, 16), 1.00), \  ((14, 14), 1.00) \\
 day 7: ((14, 15), 1.00), \  ((14, 14), 1.00) \\
 day 8:  ((14, 14), 1.00) \\  
 ... \\
 day 12: ((14, 14), 0.80) \\
 day 13: ((14, 14), 0.97), \  ((15, 14), 0.96) \\
 day 14: ((14, 14), 0.96), \  ((15, 14), 0.95) \\
 ... \\  \hline
\end{tabular}
  \vspace*{.2cm}
  \caption{Locations and food probabilities at the beginning of each day
    (on a run on 15x15, 0.02 motion noise). The real food is on a single
    location with true coordinates of (14, 14) (see \sec \ref{sec:exps_food_stats}).
  }
\label{tab:goals}
\end{table}

With a barrier portion of 0.3 and change-rate of 0.1 on 15x15 and 0
motion noise, the memory type <1, \barr> (\ie \barrb was seen
yesterday at a location $x$), at end of day 2, predicts \barrb (at
same location $x$) with probabilities that are mostly 1.0 and
sometimes 0.8 (expected 0.9, and the remainder probability going to
\empt).  By end of day 20, these probabilities show a larger numeric
diversity, but, with more experience, less variance and closer to 0.9
(a few runs: 1.0, 0.76, .80, 0.92, ...). And <2, \barr> (from two days
ago) has a lower precision (probability for \barr) as expected
(expected 0.81), at day 20 (a few runs): 1.0, 0.52, .56, 1.0, 0.8, ...
With a higher change-rate of 0.3, at end of day 2, a few probabilities
for \barr, given by <1, \barr>, are: 0.4, 1.0, 1.0, 0.2, ..., and at
end of day 20: 0.64, 0.8, 0.6, 0.72, ... (expected 0.7). Similarly for
<1, \empt> (which predict \empt with a higher probability, and other
memory types such as for food. And as we add motion noise, the
imprecision grows.  Note that these probabilities are not only a
function of how many samples have been observed, but also the biased
towards the paths that the agent takes.

\section{Related Work }
\label{sec:related}

Our work is related to diverse tracks of research on learning and
decision making, including
the nature and use of memory (in biological and artificial systems),
reinforcement learning (RL), planning, cognitive architectures,
autonomous agents, and robotics. We focus on closely related work that
was not discussed earlier.

Two broad behaviors or strategies have been identified in
computational neuroscience: model-based (\eg map making, and goal
oriented) \vs habitual instrumental behavior (corresponding to our
\pathb strategy and typical model-free RL):
The evidence and the relative strengths and weaknesses are discussed
in \cite{Daw2018AreTwo} (such as the simplicity of the habitual \vs
the flexibility but the computational requirements of goal-oriented
model-based behavior).  Human memory representations are complex and
are sufficiently flexible to have a diversity of uses, \eg not just
for spatial maps, but, for instance, also for the more general
cognitive graphs (diverse uses)
\cite{structuringKW2020,Reagh2023FlexibleRO}. Unsupervised learning of
(explicit) structured representations, that would find repeated
diverse use, may also be foundational for perception \cite{pgs5}.


In a partially observed (limited sensing) task \cite{merlin2018}, the
authors study how (predictive) memory and RL techniques could be put
together in a perceptually realistic navigation setting, showing
promising results that the agent using predictive memory was
substantially more successful than plain model-free RL agents (\eg
remembering and finding the way to goal when teleported), but the
number of episodes (environment steps) remained considerable (\eg 100s
of thousand or millions) and generalization ability remains unclear.
In a recent study \cite{vafa2024evaluating}, the authors tackle the
tricky question of LLM generalization, and develop novel evaluation
techniques (\eg querying under small perturbations) for a few tasks.
For a navigation task (in Manhattan, NY), they present evidence
that the (transformer) generative network does not learn a systematic
('coherent') map from its sequence training data, \eg not generalizing
as well to sequences that are perturbed from the shortest paths
sequences used for training.\footnote{The perturbation to action
sequences, by which the authors study network performance in one of
their experiments, is akin to our random barrier location changes, and
to a lesser extent to our motion noise: in their work, the agent (the
transformer) is given the alternative move taken, and is queried for a
plausible next move. }

Recent work, motivated by the possibility of map construction by
humans, experiments on subjects as well as performs computational
modeling providing evidence that humans build map structures
consistent with a Bayesian approach \cite{Sharma2021MapIC},
and
furthermore, such maps help planning via partially observed Markov
decision processes (POMDPs), in a continual (re)planning fashion.  Our
study focuses on how repeated change and uncertainty could help or
limit the benefits of learning a map,
in a simple daily agentic foraging task, and the environment and interaction
durations are parameterized (attributes such as grid size and
uncertainty characteristics can be changed substantially) and we
compare a variety of strategies (pure-sensing or smell-based, and
path-memory techniques).

Change, in machine learning and RL, continues to be studied and remains
a challenge,
and techniques in  areas such as continual learning,
transfer learning, distribution shift, meta learning, lifelong
learning, lifelong RL, and open-ended learning in robotics attempt to
address the various
dimensions of the problem
\cite{Thrun1993LifelongRL,Schaul2018TheB2,Parisi2018ContinualLL,botvinick2019reinforcement,Sutton2022TheAP,Santucci2020EditorialIM}.
%
The issue of environment change facing animals was also studied in
\cite{zhang2015ReinforcementLA} and the authors propose that animals
achieve change detection and change of strategy in part through
counter-factual reasoning (as typical RL does not completely explain
the observed speed of change in behavior). See also work on replay
\cite{Foster2017ReplayCO}. In the area of simultaneous localization
and mapping (SLAM) for robotics, richer perception and change is also
identified as a future area of research \cite{slam2016}.

The work on cognitive architectures explores how components such as
perception, memory, and control can be combined, inspired by research
in human and animal intelligence, with the goal of shedding further
insight onto human intelligence as well as striving to create general
problem solvers with applications to areas such as robotics
\cite{Kotseruba201840YO,2016ALC,Langley2009CognitiveAR}. Our work
began with a more narrow focus (memory and navigation), but
consideration of different environments and somewhat different tasks
(Search \vs Plan) also extended our final solution to be somewhat more
general than originally anticipated.
Our work also shares goals with research on artificial animals,
so-called {\em animats}, in that we study intelligence in the context
of an
agent with limited sensing with the goal of sustaining itself
\cite{wilson1985KnowledgeGI,Strannegrd2018LearningAD,Strannengard2024}.
In particular, Stranneg{\aa}rd \etal advocate for agent architectures
and learning mechanisms for a diversity of situations (worlds and
agent needs, supporting multiple objectives, and different
sensing/motor capabilities), and explore dynamically growing memory
(learning) structures \cite{Strannegrd2018LearningAD}. In that work,
the patterns learned are deterministic and the feedback (reward/cost)
was not delayed significantly. We focused on navigation under change
and uncertainty as well as the use of (episodic) memory.

\section{Summary and Future Directions }
\label{sec:summary}


Organisms have
limited access to external worlds that are uncertain, complex, and
non-stationary. This includes changes in the environment, \eg in the
weather (temporary, periodic, or permanent change) as well as in one's
capabilities (\eg losing an arm, or gaining new skills). We focused on
a closed but changing world, and on developing agents that can
remember and learn fast (learning along a life trajectory, keeping
pace with change), and accepted
that certain aspects of the strategies that the agent deploys are
hardwired (\eg geared towards the navigation task), while other aspects
could be learned and tuned continually.  We showed that an agent
with significant memory and computing (planning), with appropriate
algorithms and architecture, can substantially out-perform a
greedy-smell agent (purely sense-based with no episodic memory) as
long as change and uncertainties are not too large.

We touched on a variety of future directions throughout the paper.  In
general,
we hope to reduce the number and extent of the 'hardwired' assumptions
we made, towards more autonomy. For instance, how could an agent come
to know what to remember (and how to use it), and how does it 'carve'
and granularize its spatial and temporal inputs? Related to this is
support for richer perception as well as {\em open worlds},
such as learning and using various environmental and task regularities
in an {\em agent friendly} manner, \ie sample efficient and robust to
limited and biased experience, continually adapting, and cumulative
when possible.

{\bf Acknowledgments.} Many thanks to the fellow members of our weekly
SAIRG reading group, in particular Sean Kugele, Georgi Georgiev,
Christian Martinez, and Gene Lewis, for valuable discussions and
pointers.

\bibliographystyle{plain}  
\bibliography{global}

\vspace*{3in}


\co{
\appendix


\section{Goal and Memory-Based Levy Walks}

Implementing Levy walks with a memory.

Levy-steps


\section{Onto the Real (Open-Ended) World}

This is to 'save' the model-based approach to decision making from
becoming irrelevant, on the path to approximating human-level
intelligence and in particular human-level decision making ...

The model-based approach to decision making problems, eg modeling the
world as MDPs or the world-agent interaction as POMDPs ...

from fully observable to partially observable to unobservable..

partially observable is more realistic in modeling agent/decision
making problems, but the partially observable model-based approach
lends itself to even harder/intractable problems.. \cite{} whenever we
want to tackle 'realistic problems'..

Is the 'model-based' approach doomed ?? (  in reaching human-level
agentic capabilities   )

Model-free approaches for RL, and beyond that for decision making in
general, mapping sensory input to actions, (such as neural nets) have
this advantage that they dont assume much.. in particular they don't
assume a world that can be modelled.. (and arguably have had much
bigger impact so far \cite{}).. but they assume too little! And can be
slow to react (to contingencies not seen in training) or solve new
related problems...  (many have raised these issues, or the advantages
of modeling, and we discuss some such in this paper too.. ) ..

Early on (more than 20 years ago), after working some time in the
field and thinking about the decision making problems, I'd say to
myself but humans don't do it this way (the pure model-based/planning
way) (too much planning.. too much modeling of a too complex of a
world!).. ( discussing with coleagues, many have had such feelings,
and the tradeoff between solving a well-defined and established or
popular task, \vs less established|ill-defined but possibly, in some
ways, more aligned with what humans do!)...

If you refer to the problem as ``partially observable'', what does
'partial observability' refer to ?? if you say my {\em \bf simulated}
environment of the agent is definitely finite, and the agent at any
time senses part of this finite environment (I suspect this is the
common implicit thought), and that's my end goal (to succeed in that
kind of environment), fine (or the ultimate application is some
factory floor, or something equivalently confined with finite number
of parts relevant to the agent's task.. but there in the real-world,
always the dangers of open-endedness is lurking..  some problems such
as driver-less cars may not be confined or closed-world... they are
likely much more open-ended than we may initially think...

( .. is this a productivity/systematicity debate too ?? )

And if the goal is to (strive to) reach those (human-level)
capabilities, then calling the problem (or the world) ``partially
observable'' is a misnomer..  is harmful..  

In a model-based approach, when you say the problem/task is partially
observable, perhaps you are hoping that agents, by viewing a few parts
of this puzzle at a time, can peace the parts together, and compute
and complete the full model of the world at some point in their
lifetime.. (.. like solving a 1000 parts puzzle .. )

However, there is still hope to save some version of (or
versions/extensions of) the 'model-based approach to decision making'.

(I believe it can lead to waste of energy? why write this article?!! )

Implicit in the phrase 'partially observable' are one of these
assumptions:
%
\begin{itemize}
\item The world has a finite number of parts, and the agent observes a
  subset of those  parts.. \footnote{Could the world have infinite
  parts, and we are observing finitely many subset..}
\item the world (composed of many parts) is out there, which certain
  organization of objects/parts, and the agent is observing parts of
  that organization, hence 'partially observable' ..
\end{itemize}

Both interpretations can be wrong/misleading.  Many philosophers have
raised the issues with assuming there is a single objective world out
there, a single (unifying, etc) Truth with a captial T!

But this does not mean there are no regualrities or structures to
capture/model.  There can be rich regularities in the sensory stream
of an agent (othewise, minds and science would not be be possible!).

Here's a view or a starting point (that I think is more fruitful):

\begin{itemize}
\item The world is {\em \bf infinite}.
\item The agent has {\em \bf finite} computational means (and related finiteness constraints).
\item (.. the agent has a goal(s) such as living a good life.. )
\item The agent repeatedly interacts with the world. This repeated
  interaction leads to a sensorimotor input stream, including rewards
  (energy/food obtained, etc).
\end{itemize}

More on the agent: finite time to make a decision and act, and of
course, finite sensors).\footnote{All the computing machinery of the
agent, at any given time point in its lifetime, is finite. But certain
aspects such as memory could grow over time.  } Thus we have a finite
agent (system, organism) interfacing an infinite world, repeatedly.

{\em Intelligence arises from this (goal-directed) interaction.}

How could modeling, the {\em model-based} approach to RL and decision
making, still make sense and remain meaningful?  What the agent
observes has much structure, and discovering such can be highly
beneficial.  In other worlds, the {\em \bf sensory stream of the
  agent, has much useful structure}! Because this stream originates
from the infinitely rich (structured but infinite) world, the
structures discovered are {\em NOT likely} to form a single cohesive
model (in an open ended application domain), but the model-based
approach, by looking for and discovering many regularities in its
input stream, can still be highly advantagoups.

So the (external) world is NOT 'partially observable', and we should
be careful when we say something like ``I am working on a partially
observed problem''.  Of course, the agent has finite sensors and can
do finite computations (this finiteness should often go without
explicitly saying).\footnote{One may say the (internal) 'world' the
agent constructs for itself is partly observed.. I don't think this
what is intended when researchers and practionaries refer to their
problem as 'partially observed' (they likely mean the world {\em out
  there} is partially observed)...  }

Conclusion: if you want to contrast with certain simulated tasks (\eg
Atari games) (with one performance criterion), a next more realistic
and more challenging step up is partial observability (vs full
observability).  But on the way to human intelligence and mind, we
need to let go of the assumption that there is a single finite
objective world out there, and we as agents, observe parts of it. We
should let go out of that assumptions for our engineered agents too.

}

\end{document}